\documentclass{article} 
\usepackage{iclr2026_conference,times}

\usepackage{amsmath,amsfonts,bm}

\def\eqref#1{equation~\ref{#1}}

\def\1{\bm{1}}

\DeclareMathAlphabet{\mathsfit}{\encodingdefault}{\sfdefault}{m}{sl}
\SetMathAlphabet{\mathsfit}{bold}{\encodingdefault}{\sfdefault}{bx}{n}

\usepackage{hyperref}
\usepackage{url}

\usepackage{tcolorbox}
\usepackage{fancyvrb,fvextra}
\usepackage{enumitem}
\usepackage{booktabs}
\usepackage{multirow}
\usepackage{subfigure}
\usepackage[fixed]{fontawesome5}
\usepackage{caption}
\usepackage{bm}
\usepackage{amsmath}
\usepackage{graphicx}
\usepackage{subcaption}
\usepackage{wrapfig}

\newcommand{\DataName}{\textsc{ASCIIEval}}
\newcommand{\DataNameTrain}{\textsc{ASCIITune}}

\title{ASCIIEval: Benchmarking Models' Visual Perception in Text Strings via ASCII Art}

\makeatletter
\def\blfootnote{\gdef\@thefnmark{}\@footnotetext}
\makeatother

\author{Qi Jia$^1$ \quad  Xiang Yue$^{3}$ \quad Shanshan Huang$^4$ \quad Ziheng Qin$^2$ \quad {Yizhu Liu}$^5$ \\[3pt] \textbf{Bill Yuchen Lin}$^6$ \quad \textbf{Yang You}$^2$\textsuperscript{\textdagger} \quad \textbf{Guangtao Zhai}$^{1,7}$\textsuperscript{\textdagger} \\[3pt]
	$^1$Shanghai Artificial Intelligence Laboratory
	\quad $^2$National University of Singapore\\[3pt]
	$^3$Carnegie Mellon University
	\quad $^4$Guangzhou University
	\quad $^5$Meituan\\[3pt]
	$^6$University of Washington
	\quad $^7$Shanghai Jiao Tong University\\[3pt]
	{\scriptsize{\faEnvelope[regular]}} \texttt{jiaqi@pjlab.org.cn}
}

\iclrfinalcopy 
\begin{document}

	\maketitle
	
	\blfootnote{\textsuperscript{\textdagger}Corresponding author.}

	\begin{abstract}

		Perceiving visual semantics embedded within consecutive characters is a crucial yet under-explored capability for both Large Language Models (LLMs) and Multi-modal Large Language Models (MLLMs). 
		In this work, we select ASCII art as a representative artifact. It depicts concepts through careful arrangement of characters, which can be formulated in both text and image modalities. We frame the problem as a recognition task, and construct a novel benchmark, ASCIIEval. It covers over 3K samples with an elaborate categorization tree, along with a training set for further enhancement.
		Encompassing a comprehensive analysis of tens of models through different input modalities, our benchmark demonstrate its multi-faceted diagnostic power. 
		Given textual input, language models shows their visual perception ability on ASCII art concepts. Proprietary models achieve over 70\% accuracy on certain categories, with GPT-5 topping the rank.
		For image inputs, we reveal that open-source MLLMs suffer from a trade-off between fine-grained text recognition and collective visual perception. They  exhibit limited generalization ability to this special kind of arts, leading to the dramatic gap of over 20.01\% accuracy compared with their proprietary counterparts.
		Another critical finding is that model performance is sensitive to the length of the ASCII art, with this sensitivity varying across input modalities. Unfortunately, none of the models could successfully benefit from the simultaneous provision of both modalities, highlighting the need for more flexible modality-fusion approaches.
		Besides, we also introduce approaches for further enhancement and discuss future directions. 
		Resources are available at \url{https://github.com/JiaQiSJTU/VisionInText}.
		
	\end{abstract}

	\section{Introduction}

While conventional wisdom suggests that texts primarily function as carriers of linguistic information and images as conveyors of visual information, real-world scenarios often involve the integration of multiple information formats. For example, images may carry textual information, thus Optical Character Recognition (OCR)~\citep{mori1992historical} has been extensively studied. It focuses on capturing and understanding linguistic information embedded in images through visual processors, which is a crucial ability required in modern models for visual reasoning tasks~\citep{liu2024ocrbench}. In contrast, the comprehension of visual information embedded within text strings has not received commensurate attention.

\begin{figure}
    \centering
    \includegraphics[width=0.96\linewidth]{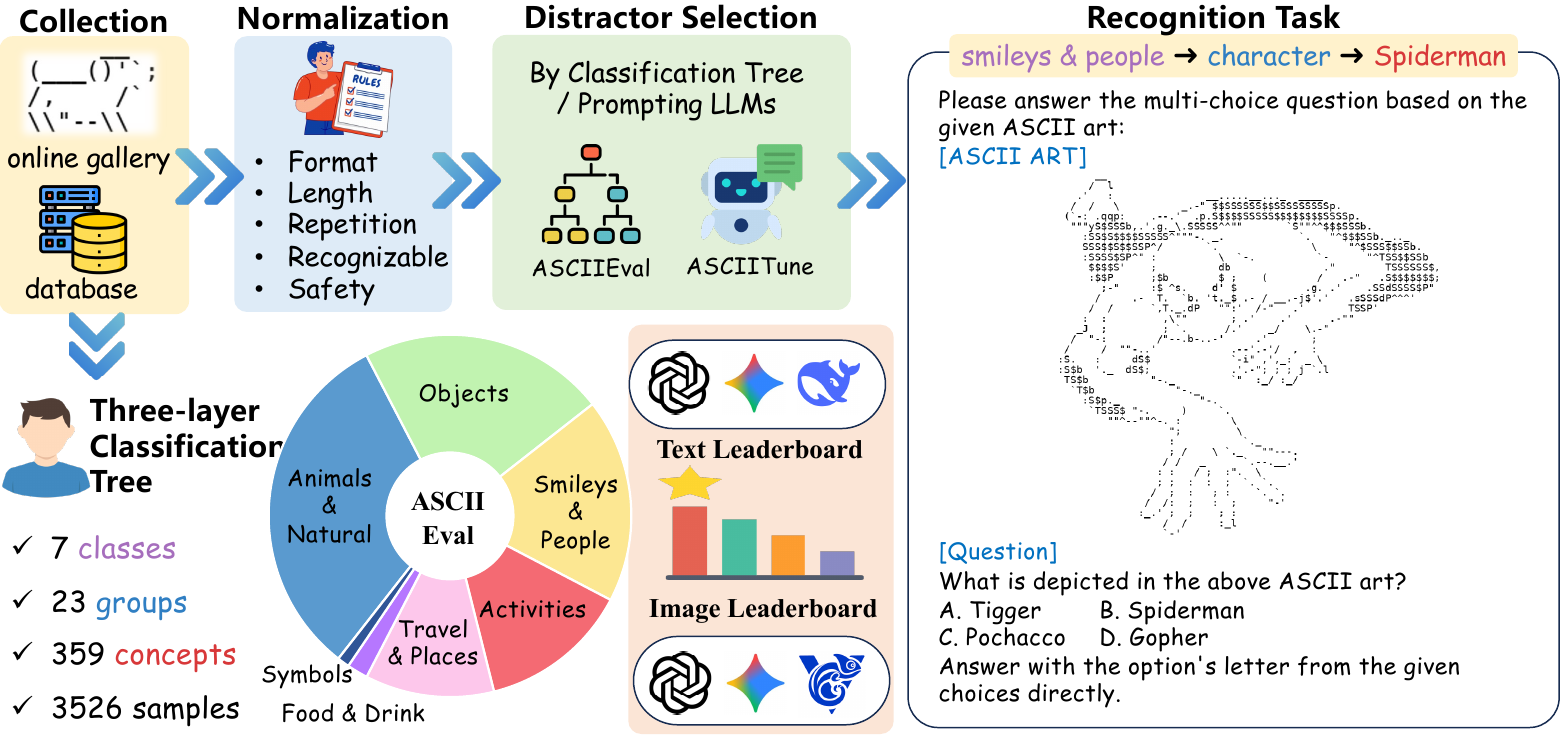}
    \caption{Overview of the \DataName{} Benchmark.}
    \label{fig:overview}
\end{figure}

Upon pre-training on a vast amount of text corpus, language models are generally hypothesized to be capable of capturing 2D structures in human writtings through escape characters, such as ``\textbackslash n''. However, they were predominately assessed via textual-semantic-based benchmarks, without focused analysis on their visual perception ability. Understanding how well models can capture visual semantics in text strings is valuable for both academic research and practical applications. A natural and representative choice is ASCII art\citep{xu2016ascii} as shown in Fig.~\ref{fig:overview}. 
Visual information in these artifacts is situated in the middle of text strings and images, and can be readily expressed in both formats containing identical content. In other words, it is modality-agnostic and therefore emerges as an ideal tool for benchmarking LLMs' visual perception ability.

As for MLLMs~\citep{achiam2023gpt,reid2024gemini,anthropic2024claude} that arm LLMs with visual processors, the character-based nature of ASCII arts presents a unique challenge. Its visual style differs starkly from images in standard benchmarks, thereby providing a rigorous test of the MLLMs' visual generalization ability. Beyond generalization, the inherent modality-agnostic quality of ASCII art serves as an excellent proxy for evaluating cross-modality alignment. A well-aligned MLLM is expected to not only perform robustly among different modalities, but also take the best of both worlds when two modalities are presented simultaneously.

Moreover, this research can also {benefit a wide range of applications and have significant safety implication for LLMs and MLLMs}. Such visual information is ubiquitous in a wide range of practical scenarios, such as processing tabular data~\citep{deng2024tables}, spatial reasoning~\citep{wu2024visualization} and playing board games~\citep{topsakal2024benchmarking}. On the safety front, using visual information reflected in characters to break through the defense line is emerging as a vulnerability for adversarial attacks~\citep{jiang2024artprompt}. For example, the attacker may use the ASCII art of a ``bomb'' instead of the word itself to circumvent safety protocols. A thorough analysis for understanding models' visual perception ability should be helpful for making proactive defense.

In this work, we investigate models' visual perception ability in text strings through ASCII arts with comprehensive evaluation and fine-tuning. 
Different from previous work that has focused on box diagrams~\citep{hayatpur2024taking,bayani2023testing}, rich-formatting texts~\citep{jiang2024artprompt}, or tone-based ASCII art~\citep{wang2023bot} that can be easily generated by rules or converted from images, we focus on ASCII art drawn by human artists, which is notably more abstract, replete with visual information, and popular among people. 
We formulate the task as a multiple-choice question-answering problem illustrated in Fig.~\ref{fig:overview}, where the answers are objective for straightforward verification. 
Then, we task models to recognize the concept depicted in the ASCII art.
Due to the lack of a dataset covering diverse categories that can thoroughly benchmark the ability of models, we collected data from different sources and cleaned manually under an elaborate categorization tree. In this way, we construct a test set dubbed \DataName{} covering 359 concepts, together with a training set with approximately 10k data points.

Our benchmark assesses over 50 proprietary models and open-source models given different modalities of ASCII Art. This set of models, featuring models released from 2023 to the present, charts the generational progress of AI systems. Our major findings are summarized as follows:

\begin{itemize}[leftmargin=0pt, nolistsep, itemindent=2em, label=$\circ$]

    \item \textbf{Language models demonstrate the ability to comprehend visual information solely from textual input.} Although performance on ASCIIEval strongly correlated with certain established benchmarks, it introduces greater challenges and reveals a widening performance gap between proprietary and open-source models. To bridge the gap, we propose rationale-assisted fine-tuning with data distilled from superior models (Sec.~\ref{sec:llm}).

    \item For image inputs, our results indicate substantial room for improvement on this straightforward recognition task. We observe a notable regression where newer-generation open-source MLLMs underperform their ancestors. Further analysis identified \textbf{a seesaw effect between OCR and ASCII art recognition: an overemphasis on improving OCR will inadvertently impair models' ability to perceive collective visual signals}. We propose two post-hoc methods for mitigation: low-resolution prompting and supervised fine-tuning (Sec.~\ref{sec:mllm}).

    \item Models exhibit different performance trends on ASCII art of increasing scale, contingent upon the input modality. When text and image information are provided simultaneously, performance degrades. This reveals an \textbf{incapacity of current models to dynamically synthesize congruent cross-modal signals}, resulting in inter-modal interference rather than synergistic enhancement (Sec.~\ref{sec:inter-modal-synergy}).

\end{itemize}

	\section{Backgrounds \& Related Work}

We present related work on LLM \& MLLM benchmarks and previous research on ASCII arts. 

\subsection{LLM \& MLLM Benchmarks}
Current LLM evaluations primarily assess capabilities in knowledge, reasoning, and instruction following through benchmarks like MMLU~\citep{hendrycks2020measuring}, Frontiermath~\cite{glazer2024frontiermath}, and Multi-IF~\cite{he2024multi}, with visual perception remaining understudied except for recent program-based approaches~\citep{qiu2024can}. Similarly, MLLM benchmarks (MMMU~\citep{yue2024mmmu}, MMStar~\citep{chen2024we}) primarily evaluate multimodal understanding using conventional images rather than text-based visual representations. These benchmarks also lack guarantees of modality equivalence in mixed inputs, which is a key characteristic of ASCII art where text and visual semantics align.

Existing ASCII-related tasks remain limited: BigBench~\citep{ghazal2013bigbench} includes basic character recognition tasks, while \citet{gu2024diverse} features only 40 varied ASCII generation samples. Current approaches often rely on automated conversions (e.g., Figlet~\footnote{\url{http://www.figlet.org/}}), risking model overfitting to transformation patterns rather than genuine visual understanding.
Differing from previous work, we focus on ASCII art depicting real-world profiles with abstract visual features. We propose ASCII recognition as foundational to generation tasks and propose \DataName{}, a dual-purpose benchmark for LLMs and MLLMs that uniquely combines semantic alignment across modalities with challenging visual abstraction.
\subsection{Research on ASCII Arts}

The origins of ASCII art date to the 1860s, evolving into a key graphic design technique as early computers utilized text characters for graphical simulation. While broadly encompassing styles like emoticons and animated art~\citep{carlsson2012future}, it strictly consists of 95 printable fixed-width ASCII characters~\citep{xu2016ascii}, ensuring cross-system consistency through textual representation.
Early research focused on ASCII art extraction from texts using byte patterns and compression analysis~\citep{hiroki2005ascii,suzuki2011text}. Later computer vision studies established two synthesis approaches: tone-based (intensity distribution) and structure-based (content outlines), with the latter proving more challenging for automation~\citep{xu2010structure,chung2022fast}.

ASCII art classification research typically converts text graphics into images, leveraging image features to enhance deep neural network accuracy~\citep{fujisawa2020ascii,matsumoto2018ascii,fujisawa2018ascii}. \cite{fujisawa2020ascii} automates ASCII art data generation to improve image classification. However, most studies rely on datasets with only five categories, limiting comprehensive analysis of LLMs' and MLLMs' visual representation capabilities.
Other works explore ASCII art for specific purposes. \cite{jiang2024artprompt} demonstrate its effectiveness in jailbreak attacks bypassing advanced defenses by representing rich-format texts as ASCII art. Conversely, \cite{wang2023bot} show that tone-based ASCII art with rich visual details is unintelligible to current LLMs, making it useful for bot detection. Additionally, \cite{wu2024visualization} use ASCII art to improve LLMs' spatial reasoning, while box diagrams—a specialized form of ASCII art—are benchmarked in tasks like recognition and generation~\citep{hayatpur2024taking,bayani2023testing}.

Our work positions ASCII art as a unique modality bridge, enabling systematic evaluation of modality-agnostic visual perception ability for both LLMs and MLLMs.
	
\section{ASCII Art Recognition}
\label{sec:data_construction}

We first define the ASCII art recognition task formally. Then, we introduced how we constructed the test and training data, dubbed \DataName{} and \DataNameTrain{}, followed by statistical analysis. 

\subsection{Problem Formulation}
\label{sec:problem_formulation}

We formulate ASCII art recognition as a multiple-choice question-answering (QA) task. Let $x_{\rm text}$ denote the raw textual representation of an ASCII art and $x_{\rm img}$ its corresponding rendered image. The model's objective is to recognize the correct concept depicted in the ASCII art from a set of candidates, $\mathcal{C}=\{c_1, c_2, ..., c_k\}$. For a Large Language Model (LLM), which processes only textual input, the prediction $\hat{y}$ is generated as follows:

\begin{equation}
\hat{y}_{\text{text}} = \operatorname{LLM}(x_{\text{text}}, \mathcal{C})
\end{equation}

A Multimodal Large Language Model (MLLM) can be prompted under two additional settings that leverage the visual modality:
\begin{align}
\hat{y}_{\text{img}} &= \operatorname{MLLM}(x_{\text{img}}, \mathcal{C}) \\
\hat{y}_{\text{multi}} &= \operatorname{MLLM}(x_{\text{img}}, x_{\text{text}}, \mathcal{C})
\end{align}

We refer to these three inference settings as Text-only, Image-only, and Text-Image, respectively. The prompt templates specified for each setting are detailed in Appendix~\ref{app:prompt_templates}.

\subsection{Dataset Construction}

We carried out the data construction process in four stages to collect a high-quality test dataset.

\textbf{Data Collection}\quad We collect ASCII art created by artists from online galleries and existing datasets.

\textbf{Classification Criteria}\quad Next, we manually designed a \textit{3-layer classification tree} after unifying the categories based on the categorical information from the original sources and removing potentially harmful categories. The most fine-grained category is named the \textbf{concept}, representing the semantic meaning reflected in the art. Similar concepts are merged into second-layer \textbf{groups}. Finally, they are grouped into seven major \textbf{classes} inspired by the iOS emoji categories.
Each concept can be depicted in various ways by artists.

\textbf{Normalization \& Filtering}\quad Subsequently, we conducted additional {filtering operations} using a combination of rules and human annotations as follows:
\begin{itemize}[leftmargin=0pt, nolistsep, itemindent=2em, label=$\circ$]
\setlength{\itemsep}{2mm}
    \item Each ASCII art string was normalized by removing redundant empty spaces at the beginning of each line and at the end of the string, without compromising its visual semantics.
    \item ASCII art consisting of more than 100 lines, not belonging to reserved categories, and repetitive to other ASCII arts under the same concept were discarded. Repetition was identified by calculating the edit distance between two ASCII strings. If the distance divided by the length of the existing string was smaller than 0.3, the new ASCII art will be considered redundant.
    \item Human annotators were tasked to filter out unrecognizable or ambiguous art, remove words in ASCII art to focus the dataset on visual perception and avoid information leakage through words, and adjust the category according to the 3-layer category tree (See more analysis in Appendix~\ref{app:data_distribution}).
\end{itemize}

\textbf{Multiple-Choice Data Construction}\quad Finally, we collected negative choices for each ASCII art by randomly sampling from other concepts within the same group. It should be noted that the ground truth labels were initially collected from the sources and subsequently verified by human annotators during the data filtering process. Each ASCII art string was then converted into an image. 

The training dataset \DataNameTrain{} is constructed in the same format requiring less human efforts. The negative choices are generated by prompting Llama-3-70B-Instruct and the unsafe samples recognized by Perspective API are filtered out. More details are shown in Appendix~\ref{sec:training_data}.

\subsection{Data Analysis}
\label{sec:data_analysis}

\begin{table}[]
    \centering
    \small
    \caption{Statistics of \DataName{} and \DataNameTrain{}. The average token count is around 300 varied for different tokenizers (See Appendix~\ref{sec:training_data}), respecting the context length limitation of all models.}
    \begin{tabular}{c|c|c|ccc|ccc}
        \toprule[1pt]
        \textbf{Dataset} & \textbf{\#Samples} & \textbf{\#Concepts} & \multicolumn{3}{c|}{\textbf{\#Characters}} & \multicolumn{3}{c}{\textbf{\#Lines}} \\
        \midrule[1pt]
        \DataName{} & 3,526 & 359 & 4 & 15,282 & 635.53 & 1 & 100 & 16.97 \\
        \DataNameTrain{} & 11,836 & 2,307 & 1 & 13,569 & 622.38 & 1 & 97 & 15.22 \\
        \bottomrule[1pt]
    \end{tabular}
    \label{tab:data_statistics}
\end{table}

As shown in Table~\ref{tab:data_statistics}, \DataName{} comprises 3,526 samples distributed across 359 concepts, 23 groups, and 7 classes. The data distribution is illustrated in Fig.~\ref{fig:overview}~(More in Appendix~\ref{app:data_distribution}). Each concept is represented by 9.82 ASCII art pieces on average, with a maximum of 170 and a minimum of 1, indicating an imbalance. \DataNameTrain{} consists of 11,836 samples across 2,307 concepts, which is more diverse but of lower quality. The number of lines in \DataName{} ranges from 1 to 100, reflecting its diversity and complexity. \DataNameTrain{} holds similar statistics. 

\textbf{Human Upper Bound}\quad We randomly extracted 100 samples from \DataName{} three times and asked three different annotators to perform the multiple-choice task. They achieved 100\%, 98\% and 97\% accuracy, respectively, demonstrating that the simplicity of this visual perception task.

	\section{Experiment Setup}

\textbf{Evaluated Models}\quad We benchmark a wide range of LLMs and MLLMs released from 2023 to 2025 from different model families. For open-source instructed models, we experiment with LLMs including \textbf{Llama}~\citep{touvron2023llama}, \textbf{Qwen}~\citep{bai2023qwen,team2024qwen2,yang2025qwen3}, \textbf{Mistral}~\citep{jiang2024mixtral}, \textbf{Gemma}~\citep{gemma_2024,team2025gemma} and \textbf{DeepSeek}~\cite{liu2024deepseek}, 
and with MLLMs containing \textbf{Llava}~\citep{liu2023llava}, \textbf{CogVLM}~\citep{wang2023cogvlm}, \textbf{Qwen-VL}~\citep{Qwen-VL,bai2025qwen2}, and \textbf{InternVL}~\cite{zhu2025internvl3}.
Besides, we selected several leading proprietary models including GPT-4o~\citep{openai2023}, GPT-5, Gemini-1.5-pro~\citep{reid2024gemini}, Gemini-2.5-pro~\cite{comanici2025gemini}, and Claude-opus-4. More in Appendix~\ref{app:eval-models}. 

\textbf{Evaluation Metrics}\quad We evaluate model performance on \DataName{} using accuracy, determined by an exact match between the model's output and the correct option. As detailed in Sec~\ref{sec:data_analysis}, the dataset exhibits a significant class imbalance across concepts. Therefore, we adopt \textit{macro-average} over each concept for quantifying model performance, and \textit{micro-accuracy} over each sample for analyzing specific ASCII art characteristics.

\section{Benchmarking Visual Perception of LLMs via ASCIIEval}
\label{sec:llm}

We first assess model performance on text inputs, then propose rationale-assisted fine-tuning to enhance LLMs' recognition ability and discuss future directions.

\begin{figure}[h!]
    \centering
    \subfigure[The Leaderboard]{
		\begin{minipage}[b]{0.35\textwidth}
		\includegraphics[width=1\textwidth]{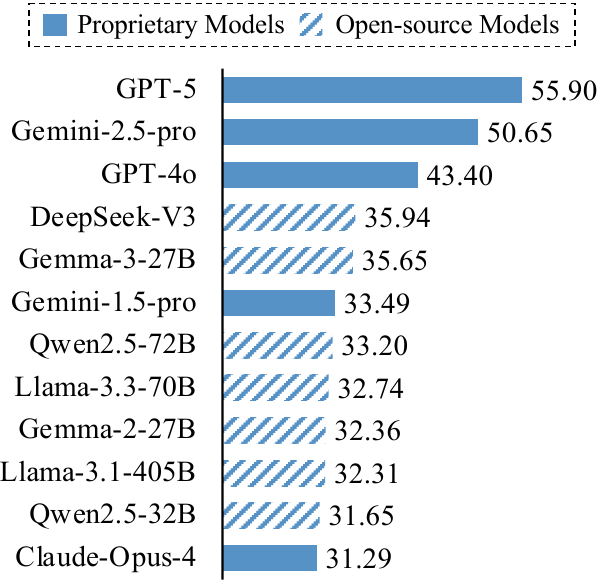}
		\end{minipage}
		\label{fig:llm_results_leaderboard}
    }
    \subfigure[Scaling Trends]{
		\begin{minipage}[b]{0.28\textwidth}
		\includegraphics[width=1\textwidth]{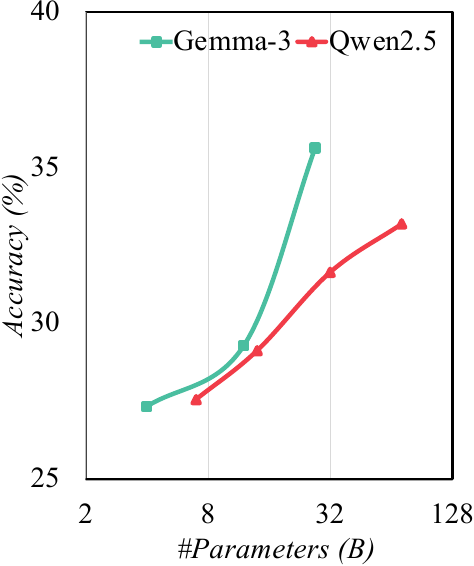}
		\end{minipage}
		\label{fig:llm_results_scaling}
    }
    \subfigure[Generational Gap]{
		\begin{minipage}[b]{0.3\textwidth}
		\includegraphics[width=1\textwidth]{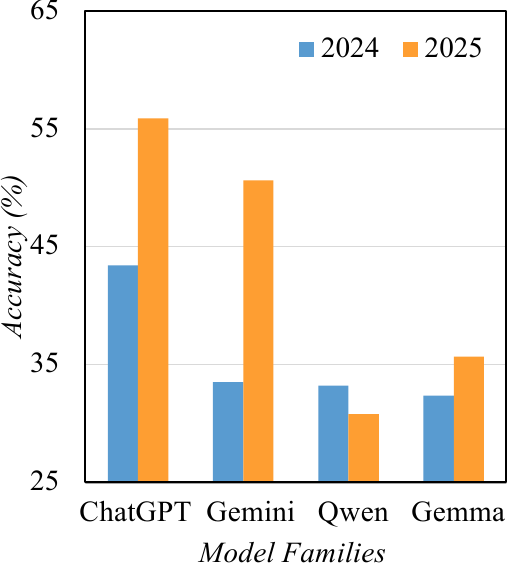}
		\end{minipage}
		\label{fig:llm_results_year}
    }
    \caption{Macro-accuracy(\%) of LLMs on \DataName{}.}
    \label{fig:llm_results}
\end{figure}

\subsection{Performance of LLMs}
\label{sec:llm_results}

Performance of LLMs and proprietary models with only text inputs is shown in Fig.~\ref{fig:llm_results}. Fig.~\ref{fig:llm_results_leaderboard} only presents a leaderboard with top-12 models. The full leaderboard is in Appendix~\ref{sec:full_leaderboard}.

\textbf{Overall Performances}\quad All of the models in Fig~\ref{fig:llm_results_leaderboard} exceeds a random baseline (25\%), {confirming their fundamental competence on visual perception through text strings}. However, a significant performance disparity exists between proprietary and open-source models. The former dominate the upper echelons of the leaderboard. The leading proprietary model, GPT-5, outperforming its open-source counterpart, DeepSeek-V3, by a substantial margin of 19.96\%. Nevertheless, all models lags far behind the human upperbound (98.33\%), reflecting the difficulty of our benchmark.

\textbf{Scaling Trends}\quad We plot the performance against the parameter count for representative models from the Gemma-3 and Qwen2.5 series in Fig~\ref{fig:llm_results_scaling}. The results indicate clear scaling trends within each single-model series. However, this scaling law does not hold across different model series. Gemma-3 with only 27B parameters outperforms other competitors with more than 70B and even hundreds of billions of parameters. This underscores the potential of developing powerful lightweight models with strong visual perception abilities.

\textbf{Generational Gap}\quad Fig~\ref{fig:llm_results_year} compares the performances of models released in 2024 with their successors from 2025 across four model families. Proprietary models indicate substantial improvements across years, with accuracy gains exceeding 10\%. In contrast, open-source models exhibit a tread of stagnation, widening the performance gap between proprietary and open-source models.

\textbf{Correlation Analysis}\quad ASCII art is not the only form of visual information embedded in text. Other representations, such as tabular data and code snippets with spatial significance, share a similar underlying requirement for this fundamental capability. To confirm this shared capability, we compared our benchmark against TableEval~\citep{zhu2025tableeval} and SGP-Bench~\citep{qiucan}, which assess LLMs on table question-answering and symbolic graphics understanding, respectively. The results show a strong positive correlation between performance on our dataset and these two benchmarks, with Pearson correlations of 0.78 and 0.85. While these findings suggest a shared fundamental skill, they also underscore the unique value of our benchmark. ASCIIEval isolates the core visual perception ability from other confounding factors such as complex reasoning, providing a more challenging and focused evaluation.

\subsection{Improving LLMs by Rationale-assisted Fine-tuning}
\label{sec:llm_improvement}

Our preliminary experiments revealed that fine-tuning LLMs on the \DataNameTrain{} by generating the choice given the multiple choice question with textual ASCII arts directly fails to yield improvements in their visual perception capabilities. Inspired by the outstanding performance of GPT-5 given the image input in Sec.~\ref{sec:mllm}, and the success of LLMs' reasoning ability by encouraging chain-of-thought, we propose rationale-assisted fine-tuning. This approach is designed to explicitly teach the model the underlying analytical process required for interpreting complex ASCII art, rather than merely exposing it to input-output pairs. It includes two primary stages as follows:

\textbf{Data Synthesis}\quad The cornerstone of our approach is the creation of a high-quality, rationale-annotated dataset. Recognizing the superior performance of state-of-the-art proprietary models, we employ GPT-5 given both $x_{\rm text}$ and $x_{\rm img}$ to synthesize the reasoning process in rich of the interpretation of local ASCII art features. 6309 instances are left after data verification.

\textbf{Rationale-assisted Fine-tuning}\quad We fine-tune the LLM on the synthesized dataset. For each instance, the model receives the original ASCII art $x_{text}$ as input. The target output is the concatenated string of the rationale and the oracle answer $y$. Further details are in the Appendix~\ref{app:rationale-assisted-details}. 

Using Qwen3-8B as the backbone, we found that both zero-shot with thinking and fine-tuning on the \DataNameTrain{} failed to improve performance, achieving 27.21\% and 26.23\% respectively. In contrast, rationale-assisted fine-tuning significantly elevated the model's accuracy from its original 28.28\% to 35.66\%, a relative gain of 26.10\%. This improvement propelled the model to fifth place on the leaderboard. Our method enabled this smaller model to outperform not only open-source models with a significantly larger number of parameters but also several proprietary models.

\subsection{Future Directions} 

\begin{wrapfigure}{r}{0.3\textwidth}
  \centering
  \includegraphics[width=0.15\textwidth]{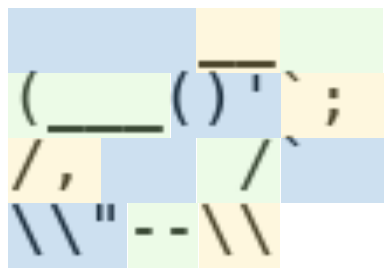}
  \caption{An illustration of the tokenized ASCII art. Each colored block represents a token.}
  \label{fig:tokenized_dog}
\end{wrapfigure}
Although Rationale-Assisted Training significantly enhances model performance, we posit that this improvement does not fundamentally enhance LLMs' ability. Its success stems from a divide-and-conquer strategy. The rationale effectively deconstructs a complex ASCII art into a series of localized sub-strings with descriptions, assisting LLMs to perform compositional reasoning at the inference time by identifying and recombining fragments memorized during training. 
We hypothesize that the bottleneck lies in the tokenization process of LLMs, which is inherently unsuitable for preserving 2D spatial information. 
For example, the dog will be processed into 13 tokens as shown in Fig~\ref{fig:tokenized_dog}. Consecutive characters will be concatenated arbitrarily,  which inevitably destroys the crucial vertical coherence of the art.  
Therefore, exploring alternative input representations is a vital furture direction.

\section{Benchmarking and enhancing MLLMs on ASCIIEval}
\label{sec:mllm}

We evaluate models on image inputs, introduce two strategies and also discuss future directions. More analysis on MLLMs' sensitivity to minor character changes and fonts is in Appendix~\ref{app:character-changes} and \ref{app:different-fonts}. 

\subsection{Performance of MLLMs}
\label{sec:mllm_results}

\begin{figure}
    \centering
    \includegraphics[width=0.85\linewidth]{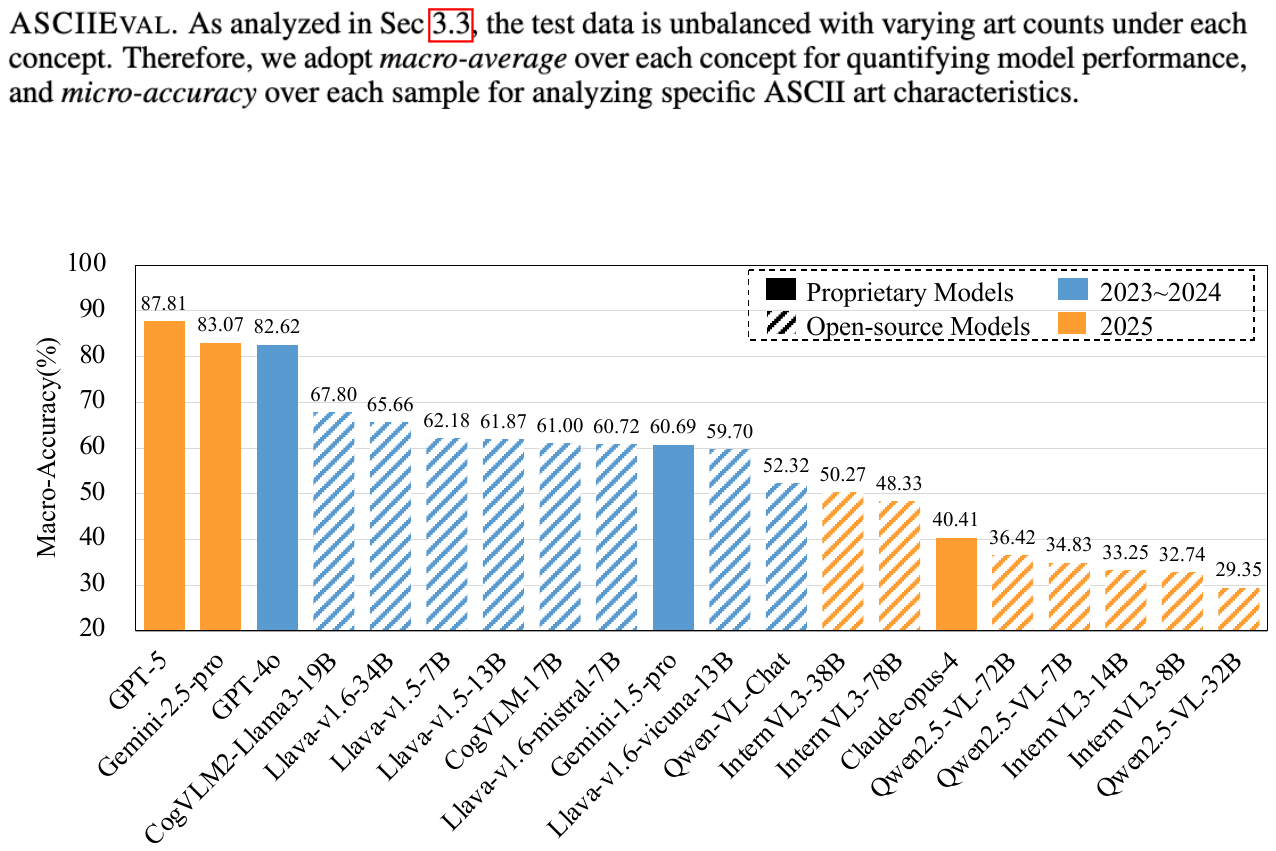}
    \caption{Macro-accuracy(\%) of MLLMs on ASCIIEVAL.}
    \label{fig:mllm_results}
\end{figure}

\textbf{Overall Performance}\quad Our evaluation reveals a clear performance hierarchy among contemporary MLLMs. At the apex of the leaderboard, proprietary models demonstrate superior capabilities, with GPT-5 achieving the highest accuracy of 87.81\%, closely followed by Gemini-2.5-pro. The top-performing open-source model, CogVLM2, attains a respectable accuracy of 67.80\% despite its relatively modest 19B parameter count.
Nevertheless, a substantial performance gap persists between the two ecosystems. GPT-5 outperforms the leading open-source model by a significant margin of 20.01\%, underscoring the current dominance of proprietary models on this visual perception task.

\textbf{Generational Gap}\quad 
A longitudinal analysis comparing models released in 2023 to 2024 with their 2025 successors highlights a diverging trend in development. Proprietary models exhibit significant year-over-year improvement, indicating a rapid advancement in their ability to interpret the abstract, symbolic nature of text strings. For instance, the Gemini family's accuracy surged from 60.69\% to 82.62\%. 
In contrast, the open-source models exhibit a marked decline in performance. Taking the Qwen-VL family as an example, the earlier model achieves 52.32\% accuracy, whereas its successor with the same number of parameters only reached 34.83\%. This regression suggests that the focus of open-source model development may be shifting away from core visual interpretation capabilities.

\textbf{Correlation Analysis}\quad 
We hypothesize that the performance decline stems from an overemphasis on benchmarks that prioritize OCR and fine-grained text extraction. As a result, models are optimized to ``read'' the characters while neglecting to ``see'' the emergent visual information they collectively form.
We analyzed the correlation between open-source MLLM performance on ASCIIEval and OCR-centric benchmarks, including OCRBench~\cite{liu2024ocrbench} and TextVQA~\cite{singh2019towards}. The results in Fig.~\ref{fig:mllm_correlation} show a negative correlation, which supports our hypothesis and suggests a fundamental trade-off between visual and text recognition.

\subsection{Improving MLLMs by Low-resolution Prompting and Fine-tuning}
\label{sec:mllm_improvement}

\begin{figure}[t]
    \centering
    \begin{minipage}{0.3\textwidth}
        \centering
        \includegraphics[width=0.98\textwidth]{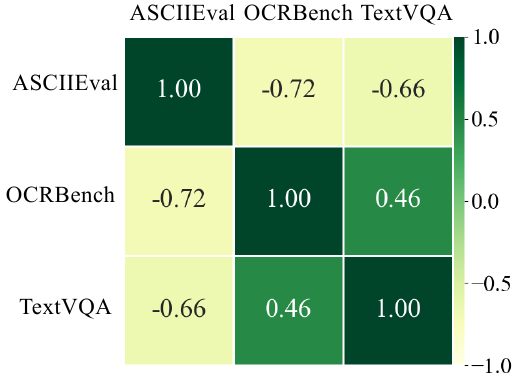}
        \captionof{figure}{Pearson Correlations between multi-modal benchmarks.}
        \label{fig:mllm_correlation}
    \end{minipage}
    \hfill
    \begin{minipage}{0.68\textwidth}
        \centering
        \captionof{table}{Macro-accuracy(\%) of Qwen2.5-VL-8B on \DataName{} by different approaches. The \textbf{best} and \underline{sub-optimal} results are in bold and underlined.}
        \begin{minipage}{0.40\textwidth}
            \centering
            \small     
            \textbf{(a)} Low-resolution prompting
            \begin{tabular}{l|c}
                \toprule[1pt]
                \textbf{Resolution} & \textbf{Accuracy} \\
                \midrule[1pt]
                default & 34.83 \\
                (1, 16) & \textbf{52.32} \\
                (1, 32) & \underline{47.65} \\
                (1, 64) & 40.59 \\
                (1, 128) & 38.81 \\
                \bottomrule[1pt]
            \end{tabular}
        \end{minipage}
        \hfill
        \begin{minipage}{0.58\textwidth}
            \centering
            \small
            \textbf{(b)} Fine-tuning strategies
            \begin{tabular}{l|c}
                \toprule[1pt]
                \textbf{Method} & \textbf{Accuracy} \\
                \midrule[1pt]
                zero-shot & 34.83 \\
                full-parameter fine-tuning & \textbf{75.83} \\
                LoRA \textit{w.} Image Encoder & \underline{75.48} \\
                LoRA \textit{w.} Text Backbone & 35.99 \\
                LoRA \textit{w.} Both & 74.23 \\
                \bottomrule[1pt]
            \end{tabular}
        \end{minipage}
        \label{fig:combined_table}
    \end{minipage}
\end{figure}

We explore two strategies to improve the performance of MLLMs.

\textbf{Low-resolution Prompting:} Since latest open-source MLLMs are optimized to read characters in images, we propose a test-time strategy by reducing the image resolution. In this way, we deliberately obscure specific characters and compel the model to percept global visual cues. We conduct experiments on Qwen2.5-VL-8B, which features the flexibility to accept a wide range of input resolutions. We set the minimum number of pixels to 1 and compared performance with a varying maximum number of pixels across the set $\{16, 32, 64, 128\}$.
Results in Table~\ref{fig:combined_table}(a) indicates a clear inverse correlation, with the lowest resolution yielding the highest accuracy.  The model achieved 52.32\% accuracy at the lowest resolution setting, outperforms the default baseline by 17.49\%. 
This finding challenges the common assumption that higher resolutions lead to better performance, suggesting that intentionally downscaling images to blur fine details is necessary in certain scenarios.

\textbf{Supervised Fine-tuning: } We investigate whether supervised fine-tuning can enhance the MLLM's capability for text-based visual perception. Using \DataNameTrain{}, the model was provided with the ASCII art image and trained to generate the correct textual answer. We train Qwen2.5-VL-8B with different fine-tuning strategies, including full-parameter fine-tuning, and parameter-efficient fine-tuning by low-rank adaptation (LoRA) on QKV matrices from different model components. 
As shown in Table~\ref{fig:combined_table}(b), the results highlight that fine-tuning the vision backbone plays the critical factor for performance improvements. Applying LoRA solely on the visual backbone achieves 75.48\%, nearly matching the full-parameter approach. Ultimately, this approach boosts Qwen2.5-VL-8B to the 4th place on the leaderboard, closely following strong proprietary models.

\subsection{Future Directions}

Our benchmark, ASCIIEval, highlights a critical but overlooked dimension of visual intelligence: holistic visual understanding. It reveals a fundamental trade-off, showing that {an overemphasis on fine-grained text recognition can come at the expense of a model's ability to perceive collective visual information}. While our proposed methods including low-resolution prompting and supervised fine-tuning, efficiently improve ASCII art performance without compromising the base model's core capabilities, they are merely pos-hoc solutions. Developing models that can intrinsically balance these competing skills is crucial for achieving the robust, state-of-the-art performance observed in leading proprietary models and for complicated real applications.

\section{The Absence of Inter-modal Synergy in MLLMs}
\label{sec:inter-modal-synergy}

\begin{figure*}[ht!]
\centering
\subfigure[\small Text-only\label{fig:line-num-text}]{
\begin{minipage}[t]{0.3\linewidth}
\centering
\includegraphics[width=\linewidth]{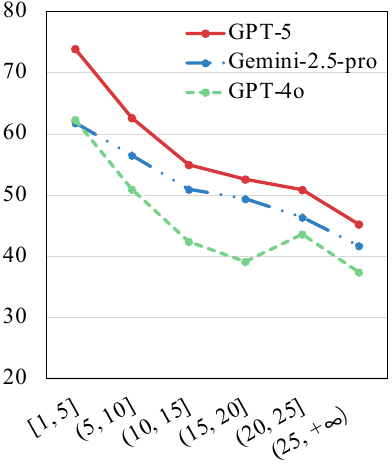}
\end{minipage}%
}%
\subfigure[\small{Image-only}\label{fig:line-num-image}]{
\begin{minipage}[t]{0.3\linewidth}
\centering
\includegraphics[width=\linewidth]{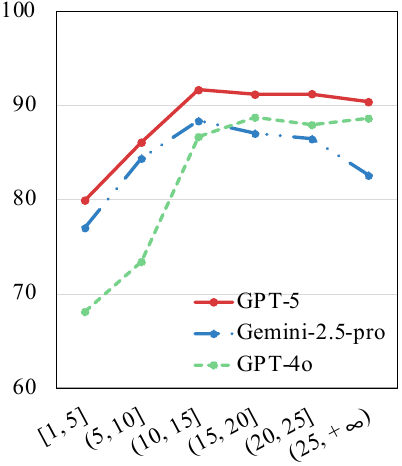}
\end{minipage}%
}%
\subfigure[\small{Comparisons}\label{fig:comparison-settings}]{
\begin{minipage}[t]{0.33\linewidth}
\centering
\includegraphics[width=0.98\linewidth]{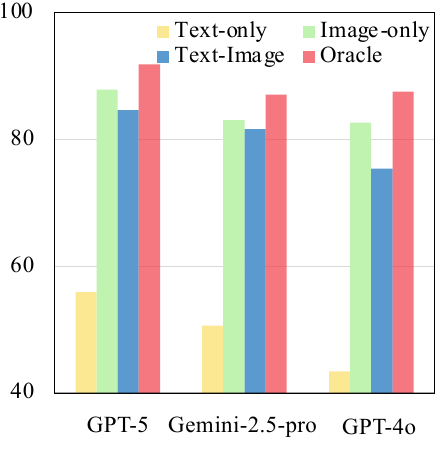}
\label{fig:compasirions_oracle}
\end{minipage}%
}%
\caption{Micro-accuracy (\%) of models on ASCII art with different numbers of characters.}
\label{fig:line_num_trend}
\end{figure*}

To investigate how the complexity of ASCII art influences model performance, we analyzed test samples partitioned into six subsets based on their line count, with the results shown in Fig.\ref{fig:line_num_trend}. 
{Models under the text-only setting demonstrate a proficiency in recognizing shorter ASCII art}, where significant features are often densely packed within consecutive characters. For instance, the string ``() \textquotesingle \textasciigrave ;'' concisely captures key features of a dog (Fig.\ref{fig:overview}), suggesting that LLMs excel at associating concepts with these dense, local character patterns. However, as the size increases, these localized features become diluted, demanding a stronger 2D perceptual ability that text-only models inherently lack. Conversely, {models given the image inputs are more adept at interpreting larger ASCII art}. This is because smaller, more abstract pieces bear little resemblance to their training data, whereas larger creations are structurally similar to real images and posters they were trained on, sharing comparable outlines and luminance contrasts, as seen with the Spiderman in Fig.~\ref{fig:overview}. 

A key finding across our experiments is a consistent performance hierarchy: Image-only $>$ Text-Image $>$ Text-only. We introduce ``oracle'' in Fig.~\ref{fig:compasirions_oracle} as a performance ceiling, which deems a prediction correct if the model succeeds with either modality alone. Our results reveals that: {the inclusion of textual information alongside the image consistently impairs model performance, rather than enabling it to approach this upper bound}. Specifically, all models exhibited a performance drop in the Text-Image setting compared to the Image-only baseline, with this degradation reaching up to 12.23\%. This exposes a fundamental weakness that, instead of effectively leveraging the complementarity and consistency between visual and textual data, current MLLMs appear to be confounded by the concurrent inputs, leading to a higher error rate.

\textbf{Future Directions:} The demonstrated failure of modality fusion presents a critical area for future research. 
Future research should therefore prioritize elucidating the internal mechanisms of modal conflict while developing architectures capable of dynamic fusion. Achieving this is a crucial step toward building robust models that flexibly synthesize all available information for a more holistic and accurate understanding.

	\section{Conclusion}

In this work, we focus on analyzing and eliciting models' visual perception ability in text strings via ASCII arts. We introduce the ASCII art recognition problem, which task models to recognize the concepts depicted by the art conveyed through different modalities. We constructed both test and training data, and conducted comprehensive evaluations with dozens of LLMs and MLLMs followed by multiple enhancement approaches. Results pinpoint that our benchmark serves as a more challenging for benchmarking LLMs' visual perception ability and MLLMs' holistic visual understanding ability. It also reveal a lack of effective fusion techniques for semantic-equivalent information across different modalities, highlighting multiple future directions.

	\bibliography{iclr2026_conference}

\begin{thebibliography}{50}
\providecommand{\natexlab}[1]{#1}
\providecommand{\url}[1]{\texttt{#1}}
\expandafter\ifx\csname urlstyle\endcsname\relax
  \providecommand{\doi}[1]{doi: #1}\else
  \providecommand{\doi}{doi: \begingroup \urlstyle{rm}\Url}\fi

\bibitem[Achiam et~al.(2023)Achiam, Adler, Agarwal, Ahmad, Akkaya, Aleman,
  Almeida, Altenschmidt, Altman, Anadkat, et~al.]{achiam2023gpt}
Josh Achiam, Steven Adler, Sandhini Agarwal, Lama Ahmad, Ilge Akkaya,
  Florencia~Leoni Aleman, Diogo Almeida, Janko Altenschmidt, Sam Altman,
  Shyamal Anadkat, et~al.
\newblock Gpt-4 technical report.
\newblock \emph{arXiv preprint arXiv:2303.08774}, 2023.

\bibitem[Anthropic(2024)]{anthropic2024claude}
Anthropic.
\newblock The claude 3 model family: Opus, sonnet, haiku, 2024.
\newblock URL
  \url{https://www-cdn.anthropic.com/de8ba9b01c9ab7cbabf5c33b80b7bbc618857627/Model_Card_Claude_3.pdf}.

\bibitem[Bai et~al.(2023{\natexlab{a}})Bai, Bai, Chu, Cui, Dang, Deng, Fan, Ge,
  Han, Huang, et~al.]{bai2023qwen}
Jinze Bai, Shuai Bai, Yunfei Chu, Zeyu Cui, Kai Dang, Xiaodong Deng, Yang Fan,
  Wenbin Ge, Yu~Han, Fei Huang, et~al.
\newblock Qwen technical report.
\newblock \emph{arXiv preprint arXiv:2309.16609}, 2023{\natexlab{a}}.

\bibitem[Bai et~al.(2023{\natexlab{b}})Bai, Bai, Yang, Wang, Tan, Wang, Lin,
  Zhou, and Zhou]{Qwen-VL}
Jinze Bai, Shuai Bai, Shusheng Yang, Shijie Wang, Sinan Tan, Peng Wang, Junyang
  Lin, Chang Zhou, and Jingren Zhou.
\newblock Qwen-vl: A versatile vision-language model for understanding,
  localization, text reading, and beyond.
\newblock \emph{arXiv preprint arXiv:2308.12966}, 2023{\natexlab{b}}.

\bibitem[Bai et~al.(2025)Bai, Chen, Liu, Wang, Ge, Song, Dang, Wang, Wang,
  Tang, et~al.]{bai2025qwen2}
Shuai Bai, Keqin Chen, Xuejing Liu, Jialin Wang, Wenbin Ge, Sibo Song, Kai
  Dang, Peng Wang, Shijie Wang, Jun Tang, et~al.
\newblock Qwen2. 5-vl technical report.
\newblock \emph{arXiv preprint arXiv:2502.13923}, 2025.

\bibitem[Bayani(2024)]{bayani2023testing}
David Bayani.
\newblock Testing the depth of chatgpt’s comprehension via cross-modal tasks
  based on ascii-art: Gpt3. 5’s abilities in regard to recognizing and
  generating ascii-art are not totally lacking.
\newblock In \emph{Findings of the Association for Computational Linguistics:
  EACL 2024}, pp.\  2063--2077, 2024.

\bibitem[Carlsson \& Miller(2012)Carlsson and Miller]{carlsson2012future}
Anders Carlsson and A~Bill Miller.
\newblock Future potentials for ascii art cac. 3, paris, france.
\newblock In \emph{Postdigital art-Proceedings of the 3rd computer art
  congress}, pp.\ ~13, 2012.

\bibitem[Chen et~al.(2024)Chen, Li, Dong, Zhang, Zang, Chen, Duan, Wang, Qiao,
  Lin, et~al.]{chen2024we}
Lin Chen, Jinsong Li, Xiaoyi Dong, Pan Zhang, Yuhang Zang, Zehui Chen, Haodong
  Duan, Jiaqi Wang, Yu~Qiao, Dahua Lin, et~al.
\newblock Are we on the right way for evaluating large vision-language models?
\newblock \emph{Advances in Neural Information Processing Systems},
  37:\penalty0 27056--27087, 2024.

\bibitem[Chung \& Kwon(2022)Chung and Kwon]{chung2022fast}
Moonjun Chung and Taesoo Kwon.
\newblock Fast text placement scheme for ascii art synthesis.
\newblock \emph{IEEE Access}, 10:\penalty0 40677--40686, 2022.

\bibitem[Comanici et~al.(2025)Comanici, Bieber, Schaekermann, Pasupat,
  Sachdeva, Dhillon, Blistein, Ram, Zhang, Rosen, et~al.]{comanici2025gemini}
Gheorghe Comanici, Eric Bieber, Mike Schaekermann, Ice Pasupat, Noveen
  Sachdeva, Inderjit Dhillon, Marcel Blistein, Ori Ram, Dan Zhang, Evan Rosen,
  et~al.
\newblock Gemini 2.5: Pushing the frontier with advanced reasoning,
  multimodality, long context, and next generation agentic capabilities.
\newblock \emph{arXiv preprint arXiv:2507.06261}, 2025.

\bibitem[Deng et~al.(2009)Deng, Dong, Socher, Li, Li, and
  Fei-Fei]{deng2009imagenet}
Jia Deng, Wei Dong, Richard Socher, Li-Jia Li, Kai Li, and Li~Fei-Fei.
\newblock Imagenet: A large-scale hierarchical image database.
\newblock In \emph{2009 IEEE conference on computer vision and pattern
  recognition}, pp.\  248--255. Ieee, 2009.

\bibitem[Deng et~al.(2024)Deng, Sun, He, Sikka, Chen, Ma, Zhang, and
  Mihalcea]{deng2024tables}
Naihao Deng, Zhenjie Sun, Ruiqi He, Aman Sikka, Yulong Chen, Lin Ma, Yue Zhang,
  and Rada Mihalcea.
\newblock Tables as texts or images: Evaluating the table reasoning ability of
  llms and mllms.
\newblock In \emph{Findings of the Association for Computational Linguistics
  ACL 2024}, pp.\  407--426, 2024.

\bibitem[Fujisawa et~al.(2018)Fujisawa, Matsumoto, Ohta, Yoshida, and
  Kita]{fujisawa2018ascii}
Akira Fujisawa, Kazuyuki Matsumoto, Kazuki Ohta, Minoru Yoshida, and Kenji
  Kita.
\newblock Ascii art category classification based on deep convolutional neural
  networks.
\newblock In \emph{2018 5th IEEE International Conference on Cloud Computing
  and Intelligence Systems (CCIS)}, pp.\  345--349. IEEE, 2018.

\bibitem[Fujisawa et~al.(2020)Fujisawa, Matsumoto, Ohta, Yoshida, and
  Kita]{fujisawa2020ascii}
Akira Fujisawa, Kazuyuki Matsumoto, Kazuki Ohta, Minoru Yoshida, and Kenji
  Kita.
\newblock Ascii art classification model by transfer learning and data
  augmentation.
\newblock In \emph{Fuzzy Systems and Data Mining VI}, pp.\  608--618. IOS
  Press, 2020.

\bibitem[Ghazal et~al.(2013)Ghazal, Rabl, Hu, Raab, Poess, Crolotte, and
  Jacobsen]{ghazal2013bigbench}
Ahmad Ghazal, Tilmann Rabl, Minqing Hu, Francois Raab, Meikel Poess, Alain
  Crolotte, and Hans-Arno Jacobsen.
\newblock Bigbench: Towards an industry standard benchmark for big data
  analytics.
\newblock In \emph{Proceedings of the 2013 ACM SIGMOD international conference
  on Management of data}, pp.\  1197--1208, 2013.

\bibitem[Glazer et~al.(2024)Glazer, Erdil, Besiroglu, Chicharro, Chen, Gunning,
  Olsson, Denain, Ho, Santos, et~al.]{glazer2024frontiermath}
Elliot Glazer, Ege Erdil, Tamay Besiroglu, Diego Chicharro, Evan Chen, Alex
  Gunning, Caroline~Falkman Olsson, Jean-Stanislas Denain, Anson Ho, Emily
  de~Oliveira Santos, et~al.
\newblock Frontiermath: A benchmark for evaluating advanced mathematical
  reasoning in ai.
\newblock \emph{arXiv preprint arXiv:2411.04872}, 2024.

\bibitem[Gu et~al.(2024)Gu, Sun, Lian, Kang, Xu, and Fan]{gu2024diverse}
Zihui Gu, Xingwu Sun, Fengzong Lian, Zhanhui Kang, Cheng-Zhong Xu, and Ju~Fan.
\newblock Diverse and fine-grained instruction-following ability exploration
  with synthetic data.
\newblock \emph{arXiv preprint arXiv:2407.03942}, 2024.

\bibitem[Hayatpur et~al.(2024)Hayatpur, Hempel, Chen, Duan, Guo, and
  Xia]{hayatpur2024taking}
Devamardeep Hayatpur, Brian Hempel, Kathy Chen, William Duan, Philip Guo, and
  Haijun Xia.
\newblock Taking ascii drawings seriously: How programmers diagram code.
\newblock In \emph{Proceedings of the CHI Conference on Human Factors in
  Computing Systems}, pp.\  1--16, 2024.

\bibitem[He et~al.(2024{\natexlab{a}})He, Li, Jang, Jia, Cao, Shah,
  Shrivastava, and Lim]{he2024ma}
Bo~He, Hengduo Li, Young~Kyun Jang, Menglin Jia, Xuefei Cao, Ashish Shah,
  Abhinav Shrivastava, and Ser-Nam Lim.
\newblock Ma-lmm: Memory-augmented large multimodal model for long-term video
  understanding.
\newblock In \emph{Proceedings of the IEEE/CVF Conference on Computer Vision
  and Pattern Recognition}, pp.\  13504--13514, 2024{\natexlab{a}}.

\bibitem[He et~al.(2024{\natexlab{b}})He, Jin, Wang, Bi, Mandyam, Zhang, Zhu,
  Li, Xu, Lv, et~al.]{he2024multi}
Yun He, Di~Jin, Chaoqi Wang, Chloe Bi, Karishma Mandyam, Hejia Zhang, Chen Zhu,
  Ning Li, Tengyu Xu, Hongjiang Lv, et~al.
\newblock Multi-if: Benchmarking llms on multi-turn and multilingual
  instructions following.
\newblock \emph{arXiv preprint arXiv:2410.15553}, 2024{\natexlab{b}}.

\bibitem[Hendrycks et~al.(2021)Hendrycks, Burns, Basart, Zou, Mazeika, Song,
  and Steinhardt]{hendrycks2020measuring}
Dan Hendrycks, Collin Burns, Steven Basart, Andy Zou, Mantas Mazeika, Dawn
  Song, and Jacob Steinhardt.
\newblock Measuring massive multitask language understanding.
\newblock In \emph{International Conference on Learning Representations}, 2021.

\bibitem[Hiroki \& Minoru(2005)Hiroki and Minoru]{hiroki2005ascii}
T~Hiroki and M~Minoru.
\newblock Ascii art pattern recognition using svm based on morphological
  analysis.
\newblock Technical report, Technical report of IEICE. PRMU 104 (670), 25--30
  (20050218), 2005.

\bibitem[Hong et~al.(2023)Hong, Zhen, Chen, Zheng, Du, Chen, and
  Gan]{hong20233d}
Yining Hong, Haoyu Zhen, Peihao Chen, Shuhong Zheng, Yilun Du, Zhenfang Chen,
  and Chuang Gan.
\newblock 3d-llm: Injecting the 3d world into large language models.
\newblock \emph{Advances in Neural Information Processing Systems},
  36:\penalty0 20482--20494, 2023.

\bibitem[Jiang et~al.(2024{\natexlab{a}})Jiang, Sablayrolles, Roux, Mensch,
  Savary, Bamford, Chaplot, Casas, Hanna, Bressand, et~al.]{jiang2024mixtral}
Albert~Q Jiang, Alexandre Sablayrolles, Antoine Roux, Arthur Mensch, Blanche
  Savary, Chris Bamford, Devendra~Singh Chaplot, Diego de~las Casas, Emma~Bou
  Hanna, Florian Bressand, et~al.
\newblock Mixtral of experts.
\newblock \emph{arXiv preprint arXiv:2401.04088}, 2024{\natexlab{a}}.

\bibitem[Jiang et~al.(2024{\natexlab{b}})Jiang, Xu, Niu, Xiang,
  Ramasubramanian, Li, and Poovendran]{jiang2024artprompt}
Fengqing Jiang, Zhangchen Xu, Luyao Niu, Zhen Xiang, Bhaskar Ramasubramanian,
  Bo~Li, and Radha Poovendran.
\newblock Artprompt: Ascii art-based jailbreak attacks against aligned llms.
\newblock In \emph{Proceedings of the 62nd Annual Meeting of the Association
  for Computational Linguistics (Volume 1: Long Papers)}, pp.\  15157--15173,
  2024{\natexlab{b}}.

\bibitem[Liu et~al.(2024{\natexlab{a}})Liu, Feng, Xue, Wang, Wu, Lu, Zhao,
  Deng, Zhang, Ruan, et~al.]{liu2024deepseek}
Aixin Liu, Bei Feng, Bing Xue, Bingxuan Wang, Bochao Wu, Chengda Lu, Chenggang
  Zhao, Chengqi Deng, Chenyu Zhang, Chong Ruan, et~al.
\newblock Deepseek-v3 technical report.
\newblock \emph{arXiv preprint arXiv:2412.19437}, 2024{\natexlab{a}}.

\bibitem[Liu et~al.(2023)Liu, Li, Wu, and Lee]{liu2023llava}
Haotian Liu, Chunyuan Li, Qingyang Wu, and Yong~Jae Lee.
\newblock Visual instruction tuning, 2023.

\bibitem[Liu et~al.(2024{\natexlab{b}})Liu, Li, Huang, Yang, Yu, Li, Yin, Liu,
  Jin, and Bai]{liu2024ocrbench}
Yuliang Liu, Zhang Li, Mingxin Huang, Biao Yang, Wenwen Yu, Chunyuan Li,
  Xu-Cheng Yin, Cheng-Lin Liu, Lianwen Jin, and Xiang Bai.
\newblock Ocrbench: on the hidden mystery of ocr in large multimodal models.
\newblock \emph{Science China Information Sciences}, 67\penalty0 (12):\penalty0
  220102, 2024{\natexlab{b}}.

\bibitem[Matsumoto et~al.(2018)Matsumoto, Fujisawa, Yoshida, and
  Kita]{matsumoto2018ascii}
Kazuyuki Matsumoto, Akira Fujisawa, Minoru Yoshida, and Kenji Kita.
\newblock Ascii art classification based on deep neural networks using image
  feature of characters.
\newblock \emph{J. Softw.}, 13\penalty0 (10):\penalty0 559--572, 2018.

\bibitem[Mori et~al.(1992)Mori, Suen, and Yamamoto]{mori1992historical}
Shunji Mori, Ching~Y Suen, and Kazuhiko Yamamoto.
\newblock Historical review of ocr research and development.
\newblock \emph{Proceedings of the IEEE}, 80\penalty0 (7):\penalty0 1029--1058,
  1992.

\bibitem[OpenAI(2023)]{openai2023}
OpenAI.
\newblock Gpt-4.
\newblock \emph{OpenAI Blog}, 2023.
\newblock URL \url{https://openai.com/research/gpt-4}.

\bibitem[Qiu et~al.(2025{\natexlab{a}})Qiu, Liu, Feng, Liu, Xiao, Collins,
  Tenenbaum, Weller, Black, and Sch{\"o}lkopf]{qiu2024can}
Zeju Qiu, Weiyang Liu, Haiwen Feng, Zhen Liu, Tim~Z. Xiao, Katherine~M.
  Collins, Joshua~B. Tenenbaum, Adrian Weller, Michael~J. Black, and Bernhard
  Sch{\"o}lkopf.
\newblock Can large language models understand symbolic graphics programs?
\newblock In \emph{The Thirteenth International Conference on Learning
  Representations}, 2025{\natexlab{a}}.

\bibitem[Qiu et~al.(2025{\natexlab{b}})Qiu, Liu, Feng, Liu, Xiao, Collins,
  Tenenbaum, Weller, Black, and Sch{\"o}lkopf]{qiucan}
Zeju Qiu, Weiyang Liu, Haiwen Feng, Zhen Liu, Tim~Z Xiao, Katherine~M Collins,
  Joshua~B Tenenbaum, Adrian Weller, Michael~J Black, and Bernhard
  Sch{\"o}lkopf.
\newblock Can large language models understand symbolic graphics programs?
\newblock In \emph{The Thirteenth International Conference on Learning
  Representations}, 2025{\natexlab{b}}.

\bibitem[Reid et~al.(2024)Reid, Savinov, Teplyashin, Lepikhin, Lillicrap,
  Alayrac, Soricut, Lazaridou, Firat, Schrittwieser, et~al.]{reid2024gemini}
Machel Reid, Nikolay Savinov, Denis Teplyashin, Dmitry Lepikhin, Timothy
  Lillicrap, Jean-baptiste Alayrac, Radu Soricut, Angeliki Lazaridou, Orhan
  Firat, Julian Schrittwieser, et~al.
\newblock Gemini 1.5: Unlocking multimodal understanding across millions of
  tokens of context.
\newblock \emph{arXiv preprint arXiv:2403.05530}, 2024.

\bibitem[Singh et~al.(2019)Singh, Natarajan, Shah, Jiang, Chen, Batra, Parikh,
  and Rohrbach]{singh2019towards}
Amanpreet Singh, Vivek Natarajan, Meet Shah, Yu~Jiang, Xinlei Chen, Dhruv
  Batra, Devi Parikh, and Marcus Rohrbach.
\newblock Towards vqa models that can read.
\newblock In \emph{Proceedings of the IEEE/CVF conference on computer vision
  and pattern recognition}, pp.\  8317--8326, 2019.

\bibitem[Suzuki(2011)]{suzuki2011text}
Tetsuya Suzuki.
\newblock Text normalization on the text art extraction method using data
  compression rate.
\newblock In \emph{Proceeding of the 17th of The Annual Meeting of the
  Association for Natural Language Processing}, 2011.

\bibitem[Team(2024{\natexlab{a}})]{gemma_2024}
Gemma Team.
\newblock Gemma.
\newblock 2024{\natexlab{a}}.
\newblock \doi{10.34740/KAGGLE/M/3301}.
\newblock URL \url{https://www.kaggle.com/m/3301}.

\bibitem[Team et~al.(2025)Team, Kamath, Ferret, Pathak, Vieillard, Merhej,
  Perrin, Matejovicova, Ram{\'e}, Rivi{\`e}re, et~al.]{team2025gemma}
Gemma Team, Aishwarya Kamath, Johan Ferret, Shreya Pathak, Nino Vieillard,
  Ramona Merhej, Sarah Perrin, Tatiana Matejovicova, Alexandre Ram{\'e},
  Morgane Rivi{\`e}re, et~al.
\newblock Gemma 3 technical report.
\newblock \emph{arXiv preprint arXiv:2503.19786}, 2025.

\bibitem[Team(2024{\natexlab{b}})]{team2024qwen2}
Qwen Team.
\newblock Qwen2 technical report.
\newblock \emph{arXiv preprint arXiv:2407.10671}, 2024{\natexlab{b}}.

\bibitem[Topsakal \& Harper(2024)Topsakal and Harper]{topsakal2024benchmarking}
Oguzhan Topsakal and Jackson~B Harper.
\newblock Benchmarking large language model (llm) performance for game playing
  via tic-tac-toe.
\newblock \emph{Electronics}, 13\penalty0 (8):\penalty0 1532, 2024.

\bibitem[Touvron et~al.(2023)Touvron, Martin, Stone, Albert, Almahairi, Babaei,
  Bashlykov, Batra, Bhargava, Bhosale, et~al.]{touvron2023llama}
Hugo Touvron, Louis Martin, Kevin Stone, Peter Albert, Amjad Almahairi, Yasmine
  Babaei, Nikolay Bashlykov, Soumya Batra, Prajjwal Bhargava, Shruti Bhosale,
  et~al.
\newblock Llama 2: Open foundation and fine-tuned chat models.
\newblock \emph{arXiv preprint arXiv:2307.09288}, 2023.

\bibitem[Wang et~al.(2024{\natexlab{a}})Wang, Luo, Wang, Yu, and
  Yan]{wang2023bot}
Hong Wang, Xuan Luo, Weizhi Wang, Melody Yu, and Xifeng Yan.
\newblock Bot or human? detecting chat{GPT} imposters with a single question.
\newblock In \emph{First Conference on Language Modeling}, 2024{\natexlab{a}}.

\bibitem[Wang et~al.(2024{\natexlab{b}})Wang, Lv, Yu, Hong, Qi, Wang, Ji, Yang,
  Zhao, Song, et~al.]{wang2023cogvlm}
Weihan Wang, Qingsong Lv, Wenmeng Yu, Wenyi Hong, Ji~Qi, Yan Wang, Junhui Ji,
  Zhuoyi Yang, Lei Zhao, Xixuan Song, et~al.
\newblock Cogvlm: visual expert for pretrained language models.
\newblock In \emph{Proceedings of the 38th International Conference on Neural
  Information Processing Systems}, pp.\  121475--121499, 2024{\natexlab{b}}.

\bibitem[Wu et~al.(2024)Wu, Mao, Zhang, Xia, Dong, Cui, and
  Wei]{wu2024visualization}
Wenshan Wu, Shaoguang Mao, Yadong Zhang, Yan Xia, Li~Dong, Lei Cui, and Furu
  Wei.
\newblock Mind's eye of llms: visualization-of-thought elicits spatial
  reasoning in large language models.
\newblock \emph{Advances in Neural Information Processing Systems},
  37:\penalty0 90277--90317, 2024.

\bibitem[Xu et~al.(2010)Xu, Zhang, and Wong]{xu2010structure}
Xuemiao Xu, Linling Zhang, and Tien-Tsin Wong.
\newblock Structure-based ascii art.
\newblock In \emph{ACM SIGGRAPH 2010 papers}, pp.\  1--10, 2010.

\bibitem[Xu et~al.(2016)Xu, Zhong, Xie, Liu, Qin, and Wong]{xu2016ascii}
Xuemiao Xu, Linyuan Zhong, Minshan Xie, Xueting Liu, Jing Qin, and Tien-Tsin
  Wong.
\newblock Ascii art synthesis from natural photographs.
\newblock \emph{IEEE Transactions on Visualization and Computer Graphics},
  23\penalty0 (8):\penalty0 1910--1923, 2016.

\bibitem[Yang et~al.(2025)Yang, Li, Yang, Zhang, Hui, Zheng, Yu, Gao, Huang,
  Lv, et~al.]{yang2025qwen3}
An~Yang, Anfeng Li, Baosong Yang, Beichen Zhang, Binyuan Hui, Bo~Zheng, Bowen
  Yu, Chang Gao, Chengen Huang, Chenxu Lv, et~al.
\newblock Qwen3 technical report.
\newblock \emph{arXiv preprint arXiv:2505.09388}, 2025.

\bibitem[Yue et~al.(2024)Yue, Ni, Zhang, Zheng, Liu, Zhang, Stevens, Jiang,
  Ren, Sun, et~al.]{yue2024mmmu}
Xiang Yue, Yuansheng Ni, Kai Zhang, Tianyu Zheng, Ruoqi Liu, Ge~Zhang, Samuel
  Stevens, Dongfu Jiang, Weiming Ren, Yuxuan Sun, et~al.
\newblock Mmmu: A massive multi-discipline multimodal understanding and
  reasoning benchmark for expert agi.
\newblock In \emph{Proceedings of the IEEE/CVF Conference on Computer Vision
  and Pattern Recognition}, pp.\  9556--9567, 2024.

\bibitem[Zhu et~al.(2025{\natexlab{a}})Zhu, Wang, Chen, Liu, Ye, Gu, Tian,
  Duan, Su, Shao, et~al.]{zhu2025internvl3}
Jinguo Zhu, Weiyun Wang, Zhe Chen, Zhaoyang Liu, Shenglong Ye, Lixin Gu, Hao
  Tian, Yuchen Duan, Weijie Su, Jie Shao, et~al.
\newblock Internvl3: Exploring advanced training and test-time recipes for
  open-source multimodal models.
\newblock \emph{arXiv preprint arXiv:2504.10479}, 2025{\natexlab{a}}.

\bibitem[Zhu et~al.(2025{\natexlab{b}})Zhu, Wang, Yu, Wu, Li, Wang, and
  Xu]{zhu2025tableeval}
Junnan Zhu, Jingyi Wang, Bohan Yu, Xiaoyu Wu, Junbo Li, Lei Wang, and Nan Xu.
\newblock Tableeval: A real-world benchmark for complex, multilingual, and
  multi-structured table question answering.
\newblock \emph{arXiv preprint arXiv:2506.03949}, 2025{\natexlab{b}}.

\end{thebibliography}
	\bibliographystyle{iclr2026_conference}
	
	\newpage
	\appendix

\section{Data License}

We express our gratitude to ASCII artists from online galleries whose fantastic creations underpin our research. In order to assess the visual perception abilities of models, we made slight modifications to the original ASCII art for the test set \DataName{}, to avoid information leakage through text hints. Meanwhile, we retained the original ASCII art and the URL to the data source. We follows the term of use guidelines from the original websites~\footnote{\url{https://asciiart.website/}, \url{https://ascii.co.uk/art}} and datasets~\footnote{\url{https://huggingface.co/datasets/apehex/ascii-art}}. Data will be released and licensed under CC BY NC 4.0, which permits only non-commercial use and is intended exclusively for research purposes.

\section{Future Directions}
\label{app:future-directions}

Based on the results and analysis, we discuss more future directions as follows:

\textbf{Constructing high-quality training data automatically.} We randomly selected 100 samples from \DataNameTrain{} for the quality check and the human annotator achieved only 70\% accuracy. This indicates that \DataNameTrain{} is much noisier than \DataName{} (98.33\%), pointing out the importance of collecting more training data with higher quality. On the one hand, utilizing the ASCII art synthesis tools to convert image datasets into ASCII art can be considered to enlarge the size of the training data, under the awareness of the style differences between the converted ones and the ones created by artists. On the other hand, more strict filtering strategies should be incorporated, such as verifying the validity of ASCII art with strong MLLMs under the Image-only setting.  
    
\textbf{Improving the model architecture.} All of the tested LLMs and MLLMs show the inability to recognize information that can be fully represented in text. One potential reason is the lack of exposure to this type of data. It may be also a result of the structural limitation of current models. As for human beings, we perceive text from the aspects of character sequences and their visual shapes at the same time, while these two aspects are conventionally distinguished into two modalities when being processed by neural models. More flexible processing techniques and architecture among modalities should not only benefit the models' visual perception ability in text strings, but also make the model closer to human beings with more efficient information processing abilities.

\textbf{Incorporating more complicated scenarios.} Currently, we only considered the basic type of ASCII art made up of 95 printable fixed-width ASCII characters. Nevertheless, there also exist more fascinating ASCII arts, such as color ASCII art, 3D ASCII art, animated ASCII art, etc. These different kinds of ASCII art are also valuable for understanding LLMs designed for video understanding~\citep{he2024ma} and 3D modeling~\citep{hong20233d}.

\section{Prompt Template}
\label{app:prompt_templates}
We adopted the following three prompt templates for different input modes:

\textit{Prompt Template for Text-only Input}

\texttt{Please answer the multi-choice question based on the given ASCII art:}\\
\\   
\texttt{[ASCII ART]}\\
\texttt{{ascii\_art}}\\
\\
\texttt{[Question]}\\
\texttt{What is depicted in the above ASCII art? \{choices\}}\\
\\
\texttt{Answer with the option's letter from the given choices directly.}\\

\textit{Prompt Template for Image-only Input}

\texttt{Please answer the multi-choice question based on the given ASCII art image.}\\
\\
\texttt{[ASCII ART]}\\
\texttt{<image>}\\
\\
\texttt{[Question]}\\
\texttt{What is depicted in the above ASCII art? \{choices\}}\\
\\
\texttt{Answer with the option's letter from the given choices directly.}

\textit{Prompt Template for Image-text Input}

\texttt{Please answer the multi-choice question based on the given ASCII art in both image and text formats.}\\
\\  
\texttt{[ASCII ART Image]}\\
\texttt{<image>}\\
\\
\texttt{[ASCII ART Text]}\\
\texttt{{ascii\_art}}\\
\\
\texttt{[Question]}\\
\texttt{What is depicted in the above ASCII art? \{choices\}}\\
\\
\texttt{Answer with the option's letter from the given choices directly.}\\

All of the models except Qwen-VL are evaluated based on these prompt templates with minor modifications to adapt to their default settings, especially for the position of the image. 

Qwen-VL is more sensitive to prompt templates according our experiments. Therefore, we adapted the above templates into Qwen-VL's original format, which is "Context: ... Question: ... Answer:".

\section{Data Collection for \DataNameTrain{}}
\label{sec:training_data}

To further elicit models' visual perception ability, 
the creation of a training set is essential. An intuitive solution is to leverage previous works on ASCII art synthesis~\citep{xu2016ascii, xu2010structure} by converting existing image datasets, such as ImageNet~\citep{deng2009imagenet}.  
A public dataset~\footnote{\url{https://huggingface.co/datasets/mrzjy/ascii_art_generation_140k}} indicates that after automatic tone-based synthesis, approximately 85\% data samples are filtered out due to poor quality. Furthermore, 
existing data conversion tools are inadequate for structure-based ASCII art, which accounts for 94\% of the data according to annotators' labels in \DataName{}. Artists also frequently combine both tone-based and structure-based features in a single artifact. 

Therefore, we chose to collect the training set in a manner similar to \DataName{} instead of relying on automatic conversion. Data sources include ASCII arts from another less organized website~\footnote{\url{https://ascii.co.uk/art}}, and the crawled content was extracted into individual ASCII art pieces based on rules derived from observations. We also included the unrecognized ASCII art that was withdrawn during the construction of \DataName{}. The normalized ASCII art is discarded if recognized as repetitive with samples in \DataName{} or among each other.

\begin{table*}[t]
    \small
    \centering
    \caption{The number of samples under each category.}
    \scalebox{0.9}{
    \begin{tabular}{c|c}
        \toprule[1pt]
         \textbf{Classes} & \textbf{Groups}  \\
        \midrule[1pt]
         animals \& natural (1,122) & animal (870), plant (130), nature (122)   \\
         
         objects (777) & object (451), electronics (192), clothing (81), furniture (53)  \\
         
         smileys \& people (644) & role (199), character (195), body (146), occupation (68), people (36)  \\
         
         activities (473) & event (207), sport (126), activity (84), instrument (35), monument (21)  \\
         
         travel \& places (406) & transportation (123), building (123), places (30)  \\
         
         food \& drink (66) & food (66)  \\
         
         symbols (38)& logo (27), astrology (11)  \\
        \bottomrule[1pt]
    \end{tabular}}
    \label{tab:data_distribution}
\end{table*}

\begin{table*}[t]
    \small
    \centering
    \caption{Statistics of token length by different tokenizers.}
    \scalebox{0.9}{
    \begin{tabular}{c|ccc|ccc}
        \toprule[1pt]
         & \multicolumn{3}{c}{\textbf{\DataName{}}} & \multicolumn{3}{c}{\textbf{\DataNameTrain{}}} \\
         & Min & Max & Avg & Min & Max & Avg \\
         \midrule[1pt]
         Llama-3 Tokenizer & 71 & 2,192 & 262.72 & 69 & 3,673 & 215.10 \\
         Mistral-v0.1 Tokenizer & 85 & 2,890 & 332.91 & 83 & 4,294 & 267.93\\
         Qwen-2 Tokenizer & 80 & 2,833 & 278.17 & 78 & 3,996 & 273.40 \\
         \bottomrule[1pt]
    \end{tabular}}
    \label{tab:token_length}
\end{table*}

Due to the large amount of data with diverse concepts, carefully categorizing data for high-quality distractors is unfeasible. Instead, we prompted Llama-3-70B-Instruct  to generate negative choices given the ground truth concept and utilized the Perspective API to filter out unsafe samples based on the concatenation of candidate choices. Samples with scores less than 0.2 across all six dimensions, i.e., toxicity, severe toxicity, identity attack, insult, profanity and threat, are retained.

\section{Data Analysis and Statistics}
\label{app:data_distribution}

During the data filtering process, we recognized that some of the ASCII art have multiple interpretations, which can be summarized into two types:

\begin{itemize}[leftmargin=0pt, nolistsep, itemindent=2em, label=$\circ$]

    \item The ASCII art itself, as a kind of art form, is abstract and ambiguous. For instance, certain depictions of cats might resemble rats. Regarding these cases, we asked human annotators to remove such unrecognizable and ambiguous art.
    \item  The ASCII art is rich in content, potentially allowing two interpretations from different aspects. For example, the third ASCII art in Fig.~\ref{fig:cases-1}, can be interpreted as a beach scene, coconut tree, sunset, etc. Most of the ASCII art in \DataName{} only contains a single object, and we also tried to remove such ambiguities by carefully designing and adjusting the classification criterion. Ultimately, there are only less than 1.67\% ambiguous cases in \DataName{}, leading to the imperfect performance of human annotators.
\end{itemize}

Finally, the number of samples and the hierarchical relationship between classes and groups of \DataName{} illustrated are shown in Table~\ref{tab:data_distribution}.

The token length of samples under the Text-only mode tokenized by three representative tokenizers is in Table~\ref{tab:token_length}. The ASCII art data used in our experiments respects the context length limitation of nowadays models.

\section{Details about Evaluated Models}
\label{app:eval-models}

For open-source instructed models, we experiment with the following LLMs and MLLMs:

\paragraph{LLMs.} \textbf{Llama}~\citep{touvron2023llama} contains three collections of generative models with different sizes, including Llama-2, Llama-3, Llama-3.1, and Llama-3.3; 
\textbf{Qwen}~\citep{bai2023qwen,team2024qwen2,yang2025qwen3} is another group of models with instructed verions, including Qwen, Qwen1.5, Qwen2, Qwen2.5 and Qwen3 series; 
\textbf{Mistral}~\citep{jiang2024mixtral} includes different versions of instruction fine-tuned models, i.e., Mistral-7B-Instruct-v0.1, v0.2 and v0.3. Besides, Mixtral-8x7B-Instruct-v0.1 and Mixtral-8x22B-Instruct-v0.1 which are pre-trained generative Sparse Mixture of Experts are also compared; 
\textbf{Gemma}~\citep{gemma_2024,team2025gemma} is a family of lightweight text-to-text models with instruction-tuned variants. We considered Gemma-2 and Gemma-3 series;
\textbf{DeepSeek}~\cite{liu2024deepseek} is a series of open-source large language models, with DeepSeek-V3 being a notable model in this family.

\paragraph{MLLMs.} \textbf{Llava}~\citep{liu2023llava} augmented a pre-trained LLM with a pre-trained vision encoder. The vision model's representations are projected into the LLM's representation space with a projection layer, and it is frozen during instruction tuning while the projector and the backbone LLM are updated; \textbf{CogVLM}~\citep{wang2023cogvlm} aims at retaining the original capabilities of the LLM while adding visual understanding abilities. Representations from the pre-trained vision transformer encoder are passed through an MLP adapter as the input, and a group of trainable visual expert modules in the attention and FFN layers are introduced into the LLM. All of the parameters except the ones from the original LLM are tuned; \textbf{Qwen-VL}~\citep{Qwen-VL} proposed a position-aware vision-language adapter for compressing image features. The model is trained through three stages, i.e., pre-training, multi-task pre-training and supervised fine-tuning; \textbf{Qwen2.5-VL}~\citep{bai2025qwen2} introduce dynamic resolution processing ad excelling in omni-document parsing; \textbf{InternVL3} consolidates language pre-training and multi-modal alignment training into a unified pre-training stage with interleaving multi-modal data. 

For proprietary models, the specific versions we evaluated are GPT-4o-20240806~\citep{openai2023}, GPT-5-20250807, Claude-opus-4-20250514, Gemini-1.5-pro~\citep{reid2024gemini} and Gemini-2.5-pro~\cite{comanici2025gemini}.

\section{\DataName{} Leaderboard}
\label{sec:full_leaderboard}

We summarize the above models performance on \DataName{} given different input modalities in Table~\ref{tab:full_leaderboard}.
The statistics used for calculating correlations in Sec.~\ref{sec:llm_results} and Sec.~\ref{sec:mllm_results} were collected by extracting scores of the overlapped models covered both in Table~\ref{tab:full_leaderboard} and corresponding leaderboards~\footnote{\url{https://github.com/wenge-research/TableEval}, \url{https://sgp-bench.github.io/}, \url{https://huggingface.co/spaces/opencompass/open_vlm_leaderboard}}. 

\begin{table}[]
    \centering
    \small
    \caption{\DataName{} Leaderboard. The scores are macro accuracy (\%) averaged among different concepts. Average refers to the mean among the three input settings if available. All of the models are ``instruct'' or ``chat'' versions. The \textbf{best} and \underline{sub-optimal} results in each group of models are in bold and underlined.}

    \begin{tabular}{l|ccc|c}
    \toprule[1pt]
       \textbf{Model}    & \textbf{Text-only} & \textbf{Image-only} & \textbf{Text-Image} & \textbf{Average} \\
    \midrule[1pt]
    \multicolumn{5}{l}{\textit{Proprietary Models}} \\
        GPT-5  & \textbf{55.90} & \textbf{87.81} & \textbf{86.40} & \textbf{76.70} \\
        GPT-4o  & 43.40 & 82.62 & 75.41 & 67.14 \\
        Gemini-2.5-pro  & \underline{50.65} & \underline{83.07} & \underline{81.64} & \underline{71.79}\\
        Gemini-1.5-pro  & 33.49 & 60.69 & 58.33 & 50.84\\
        Claude-opus-4  & 31.29 & 40.41 & 36.68 & 36.13\\
    \midrule[1pt]
    \multicolumn{5}{l}{\textit{Open-source LLMs}} \\
        DeepSeek-V3  & \textbf{35.94} & - & - & \textbf{35.94} \\
        Qwen3-8B  & 28.28 & - & - & 28.28 \\
        Qwen3-14B  & 30.79 & - & - & 30.79 \\
        Qwen3-32B  & 30.18 & - & - & 30.18 \\
        Qwen2.5-7B  & 27.57 & - & - & 27.57 \\
        Qwen2.5-14B & 29.14 & - & - & 29.14 \\
        Qwen2.5-32B & 31.65 & - & - & 31.65 \\
        Qwen2.5-72B & 33.20 & - & - & 33.20 \\
        Qwen2-7B & 27.71 & - & - & 27.71 \\
        Qwen2-72B & 30.73 & - & - & 30.73 \\
        Qwen1.5-7B & 26.71 & - & - & 26.71 \\
        Qwen1.5-110B & 30.28 & - & - & 30.28 \\
        Qwen-7B  & 23.30 & - & - & 23.30 \\
        Gemma-3-4B  & 27.34 & - & - & 27.34 \\
        Gemma-3-12B & 29.29 & - & - & 29.29 \\
        Gemma-3-27B  & \underline{35.65} & - & - & \underline{35.65} \\        
        Gemma-2-9B & 30.50 & - & - & 30.50 \\
        Gemma-2-27B & 32.36 & - & - & 32.36 \\
        Llama-3.3-70B  & 32.74 & - & - & 32.74 \\
        Llama-3.1-8B & 27.22 & - & - & 27.22 \\
        Llama-3.1-70B  & 31.27 & - & - & 31.27 \\
        Llama-3.1-405B  & 32.31 & - & - & 32.31 \\
        Llama-3-8B  & 28.71 & - & - & 28.71 \\
        Llama-3-70B  & 30.42 & - & - & 30.42 \\
        Llama-2-7B  & 24.59 & - & - & 24.59 \\
        Llama-2-13B  & 25.93 & - & - & 25.93 \\       
        Llama-2-70B  & 28.08 & - & - & 28.08 \\
        Mistral-7B-v0.1  & 26.88 & - & - & 26.88 \\
        Mistral-7B-v0.2  & 26.28 & - & - & 26.28 \\
        Mistral-7B-v0.3  & 25.57 & - & - & 25.57 \\
        Mixtral-8x7B-v0.1 & 25.31 & - & - & 25.31 \\
        Mixtral-8x22B-v0.1  & 28.20 & - & - & 28.20 \\
    \midrule[1pt]
    \multicolumn{5}{l}{\textit{Open-source MLLMs}} \\
        Qwen2.5-VL-7B  & 25.05 & 34.83 & 37.01 & 32.30\\
        Qwen2.5-VL-32B  & 29.82 & 29.35 & 32.07 & 30.41\\
        Qwen2.5-VL-72B  & \textbf{34.20} & 36.42 & 37.82 & 36.15 \\
        Qwen-VL  & 24.79 & 52.32 & 40.09 & 39.07\\
        InternVL3-8B  & 27.30 & 32.74 & 33.58 & 31.21\\
        InternVL3-14B  & 25.91 & 33.25 & 31.50 & 30.22\\
        InternVL3-38B  & 32.10 & 50.27 & 47.28 & 43.22\\
        InternVL3-78B  & \underline{33.55} & 48.33 & 48.54 & 43.37\\
        CogVLM2-Llama3-19B  & 24.73 & \textbf{67.80} & \textbf{66.68} & \textbf{53.07}\\
        CogVLM-17B  & 21.25 & 61.00 & 57.58 & 46.61 \\
        Llava-v1.6-mistral-7B & 25.89 & 60.72 & 59.02 & 48.54\\
        Llava-v1.6-vicuna-13B  & 26.03 & 59.70 & 56.55 & 47.43\\
        Llava-v1.5-7B  & 24.66 & 62.18 & \underline{61.52} & 49.45\\
        Llava-v1.5-13B  & 26.00 & 61.87 & 60.70 & 49.52\\
        Llava-v1.6-34B  & 28.62 & \underline{65.66} & 61.33 & \underline{51.87}\\
    \bottomrule[1pt]
    \end{tabular}
    \label{tab:full_leaderboard}
\end{table}

\section{Data Synthesis and Training Details}
\label{app:rationale-assisted-details}

Recognizing the superior performance of state-of-the-art proprietary models, we devised a multi-step data synthesis pipeline:
\begin{itemize}[leftmargin=0pt, nolistsep, itemindent=2em, label=$\circ$]
\setlength{\itemsep}{2mm}
    \item  \textit{Data Curation:} We first employ a high-performing open-source model to filter the ASCIITune. This initial pass serves to remove low-quality or ambiguous samples, yielding 8925 samples. 
    \item \textit{Rationale Generation:} For each curated data point, we provide the teacher model, GPT-5, with both $x_{\rm text}$ and $x_{\rm img}$. The model is prompted to first generate a detailed analytical process, i.e. rationale, that describes its the reasoning process for recognition in rich of the interpretation of local ASCII art features, with the answer at the end of the output.
    \item \textit{Fidelity Verification:} Only 6309 instances where GPT-5's final answer is correct are retained.
\end{itemize}

\begin{figure}[h]
    \centering
    \includegraphics[width=0.85\linewidth]{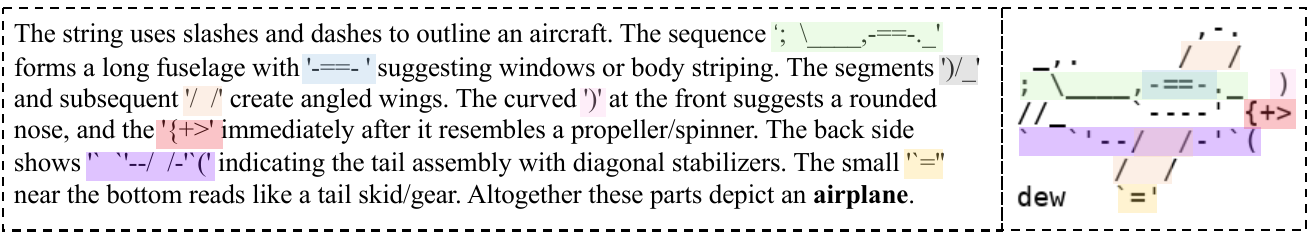}
    \caption{An example of distilled data for rationale-assisted fine-tuning.}
    \label{fig:rationale_example}
\end{figure}

An example of the output distilled from GPT-5 is shown in Fig.~\ref{fig:rationale_example}. Different colors marks the corresponding string in output text and the ASCII image. The output analysis explains the details of the model's perception process reasonably, but also contains minor errors. ``)/\_'' is a piece of hallucinated text string which not included in the original ASCII art. Employing more rigorous filtering strategies to remove such mistakes for high-quality data collection will be considered in the future.

The LLMs in Sec.~\ref{sec:llm_improvement} is finetuned on the distilled data for 2 epoch on full parameters, with batch size equaling 16 and learning rate equaling 2e-5. The MLLMs trained in Sec.~\ref{sec:mllm_improvement} adopted the same batch size and learnin rate. We did fine-tuning with full parameters for 1 epoch and with LoRA for 2 epochs.

\section{Analysis on Samples under Different ASCII Art Sizes}
\label{app:complexity}

\begin{table*}[h]
    \centering
    \small
    \caption{The number of samples with ASCII arts divided by different characteristics.}
    \begin{tabular}{c|ccccccc}
        \toprule[1pt]
         \#Characters& [1, 50] & (50,100] & (100, 200] & (200, 400] & (400, 800] & (800, 1600] & (1600, +$\infty$) \\
          \#Samples & 221 & 366 & 546 & 710 & 760 & 618 & 305 \\
        \midrule[1pt]
         \#Lines & [1,5] & (5, 10] & (10, 15] & (15, 20] & (20, 25]  & (25,+$\infty$) & -\\
         \#Samples & 414 & 854 & 699 & 534 & 399 & 626 & -\\
         \bottomrule[1pt]
    \end{tabular}
    \label{tab:data_difficulty}
\end{table*}

\begin{figure*}[ht!]
\centering
\subfigure[\small{Text-only}\label{fig:character-num-text}]{
\begin{minipage}[t]{0.49\linewidth}
\centering
\includegraphics[width=\linewidth]{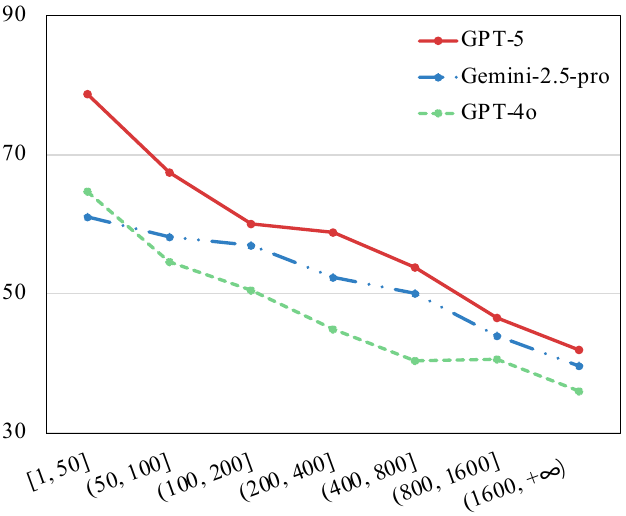}
\end{minipage}%
}%
\subfigure[\small{Image-only}\label{fig:character-num-image}]{
\begin{minipage}[t]{0.49\linewidth}
\centering
\includegraphics[width=\linewidth]{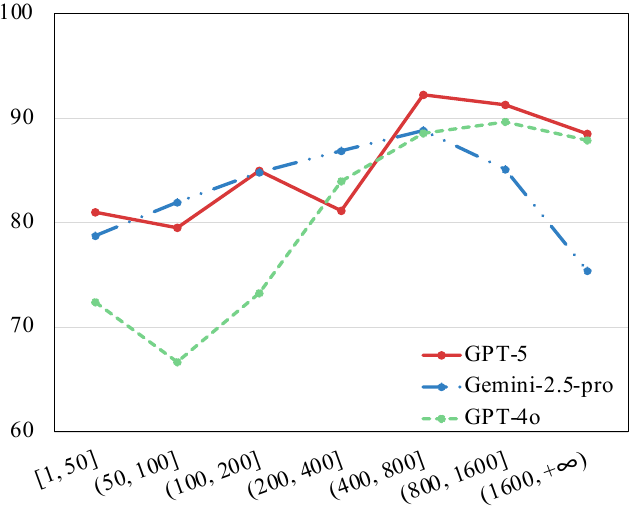}
\end{minipage}%
}%
\caption{Micro accuracy(\%) of models on recognizing ASCII arts with different numbers of characters.}
\label{fig:character_num_trend}
\end{figure*}

Based on the length characteristics of different ASCII art, we divided the test set into various subsets, as shown in Table~\ref{tab:data_difficulty}.

The performances of models on testing samples grouped by the number of lines contained in the ASCII art are shown in Fig.~\ref{fig:character_num_trend}. The trends are similar to those grouped by the number of characters in Sec~\ref{sec:inter-modal-synergy}, i.e., models favor smaller ASCII art under the Text-only setting while they prefer larger ASCII art under the Image-only setting. Besides, when an ASCII art exceeds 800 characters, the model's performance tends to plateau or even degrade, underscoring that recognizing large-scale ASCII art also remains challenging for MLLMs.

\begin{figure*}[h]
    \centering
    \includegraphics[width=0.95\linewidth]{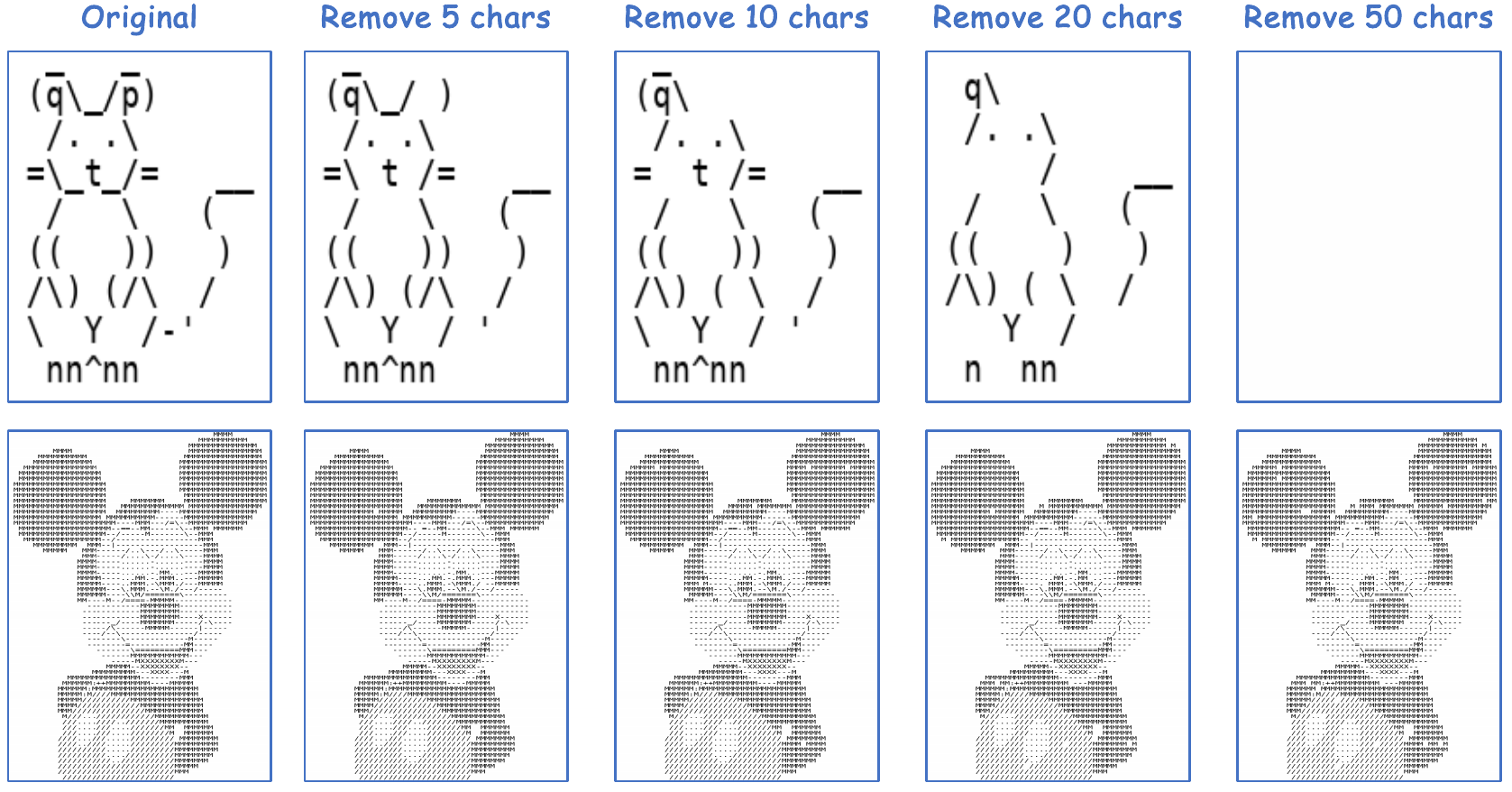}
    \caption{An illustration of removing characters in the ASCII art. ``chars'' is short for ``characters''.}
    \label{fig:remove}
\end{figure*}

\section{Sensitivity to Minor Character Changes}
\label{app:character-changes}

We randomly removed tokens (other than spaces, ``\textbackslash n'' and ``\textbackslash t'') from ASCII art and manually checked if the result remained recognizable. Two representative examples are illustrated in Fig.~\ref{fig:remove}. In both cases, the ASCII art remains recognizable when only few characters are removed. However, the first ASCII art becomes progressively indistinguishable as more characters are missing. Meanwhile, the second one just gradually has some additional noise and remains recognizable. This suggests that as the number of characters increases, the importance of each character diminishes as it carries less visual information.

We did more quantitative analysis by sampling 100 cases from \DataName{}, among which Llava-v1.6-34B provided correct answers under all three test settings. Next, we randomly replaced 1\%, 5\%, 10\%, and 20\% of tokens (other than spaces, ``\textbackslash n'' and ``\textbackslash t'') in the original ASCII art with spaces. 

The computed micro-accuracy of Llava-v1.6-34B under different test settings, as well as the human upper bound, are shown in Table~\ref{tab:perturbation}. Changing the characters in ASCII art will make the recognition task more challenging both for humans and the model, while Human is relatively more robust than Llava-v1.6-34B under different settings.

\begin{table}[h]
    \small
    \centering
    \caption{The micro-accuracy (\%) at different perturbation ratios. ``PR'' is short for ``Perturbation Ratio''.}
    \begin{tabular}{p{0.45cm}|cccc}
        \toprule[1pt]
        \text{\textbf{PR}} & \textbf{Human} & \textbf{Text-only} & \textbf{Image-only} & \textbf{Text-Image} \\
        \midrule[1pt]
        1\% & 99 & 94 & 96 & 96 \\
        5\% & 99 & 95 & 91 & 93 \\
        10\% & 97 & 91 & 93 & 92 \\
        20\% & 94 & 84 & 87 & 83 \\   
        \bottomrule[1pt]
    \end{tabular}
    \label{tab:perturbation}
\end{table}

\section{Sensitivity with Different Fonts}
\label{app:different-fonts}

\begin{figure*}[h]
    \centering
    \includegraphics[width=0.95\linewidth]{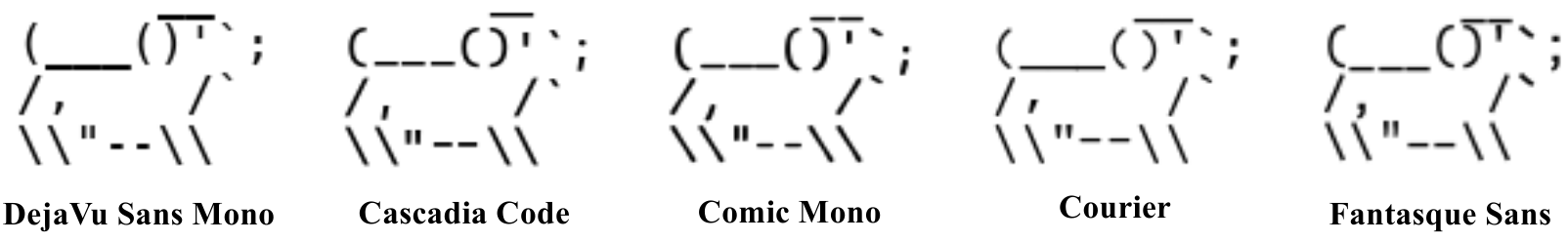}
    \caption{An illustration of an ASCII art displayed in different fixed-width fonts.}
    \label{fig:fonts}
\end{figure*}

\begin{table*}[h]
    \small
    \centering
     \caption{Macro-accuracy(\%) of Llava-v1.6-34B under Image-only and Text-Image setting with ASCII art rendered by different fix-width fonts.}
    \scalebox{1.0}{
    \begin{tabular}{c|ccccc}
        \toprule[1pt]
         \textbf{Mode} & DejaVu Sans Mono & Cascadia Code & Comic Mono & Courier & Fantasque Sans \\
        \midrule[1pt]
        \textbf{Image-only} & 65.66 & 63.41 & 66.68 & 63.84 & 66.73 \\
        \textbf{Text-Image} & 61.33 & 59.85 & 62.11 & 59.89 & 64.04 \\
        \bottomrule[1pt]
    \end{tabular}}
    \label{tab:fonts}
\end{table*}

\begin{figure*}[ht!]
\centering
\subfigure[\small Text-only\label{fig:classes-llm}]{
\begin{minipage}[t]{\linewidth}
\centering
\includegraphics[width=0.9\linewidth]{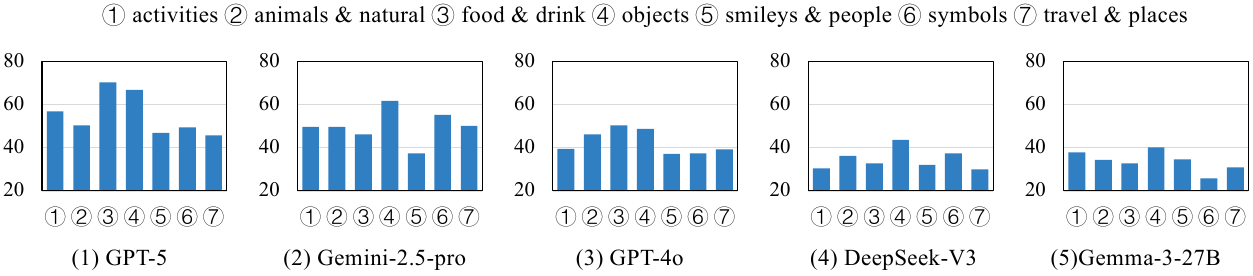}
\end{minipage}%
}%

\subfigure[\small{Image-only}\label{fig:classes-MLLM}]{
\begin{minipage}[t]{\linewidth}
\centering
\includegraphics[width=0.9\linewidth]{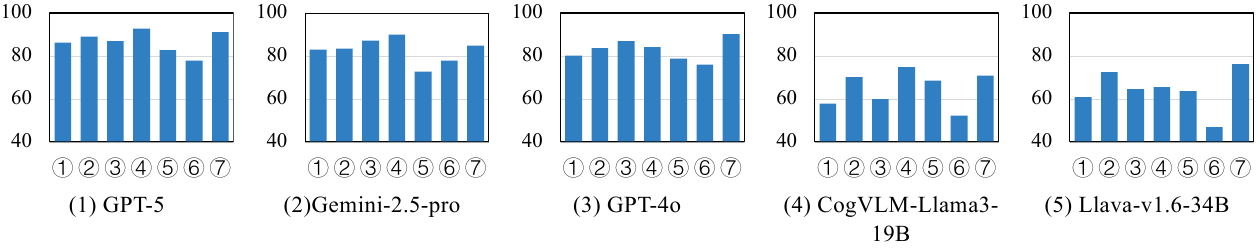}
\end{minipage}%
}%
\caption{Macro-accuracy (\%) of models on recognizing ASCII arts under different classes.}
\label{fig:classes_complexity}
\end{figure*}

In this work, we only considered the traditional ASCII art composed of 95 printable fixed-width ASCII characters. The semantic meaning remains unchanged as long as it is displayed with a fixed-width font. In addition to the ``DejaVu Sans Mono'' font used in this work, examples of the same ASCII art rendered with 4 different fonts are shown in Fig.~\ref{fig:fonts}. All of the dogs are recognizable, with only minor differences. In other words, the multiple-choice questions for ASCII art recognition in \DataName{} remain valid, regardless of the specific fixed-width font used.

Although humans have no difficulty recognizing ASCII art rendered with different fonts, this raises the question of whether MLLMs are sensitive to these variations and show a preference to a specific fixed-width font. We take Llava-v1.6-34B as an example and evaluated its performance on ASCII art under both Image-only and Text-Image settings where the images are rendered using 5 different fonts mentioned in Fig.~\ref{fig:fonts}. It should be noted that the textual ASCII art is unaffected by font variations, and Llava-v1.6-34B's performance under the Text-only setting is identical to the result in Table~\ref{tab:full_leaderboard}.

According to the results in Table~\ref{tab:fonts}, MLLMs do face challenges in performing robustly among different text fonts in ASCII art recognition and the performance varies. Nevertheless, its best performance in this table with 66.73\% and 64.04\% still lags far behind that of GPT-4o with 83.69\% and 76.52\% under both settings respectively. Moreover, the accuracy under the Text-Image setting is consistently lower than that under the Image-only setting. These observations are same as the results in Sec.~\ref{sec:inter-modal-synergy}.

On the one hand, how to reduce this sensitivity and improve the MLLMs' robustness is important and worth further exploration. On the other hand, changing the fonts in rendered ASCII art can potentially a useful data augmentation technique for boosting MLLMs' performance on \DataName{}.

\section{How do models perform on different categories?}
\label{sec:categories}

Models' performances across the 7 different classes are shown in Fig.~\ref{fig:classes_complexity}. Models given text input perform better at recognizing ASCII arts belonging to the ``objects'' class. Models under Image-only mode show consistent improvement in recognizing ``travel \& places'' over Text-only mode compared to other classes relatively. Moreover, all models struggle with ASCII art referring to ``symbols'', which comprise different logos and astrology symbols. MLLMs actually perform quite well at recognizing well-known logos, such as Apple and Linux, where GPT-5 achieves 100\% macro-accuracy and CogVLM2-Llama3-Chat-19B gets 91.16\%. However, their performance drops dramatically on relatively niche astrology symbols. Nevertheless, it is simple for both LLMs and MLLMs to answer the question ``Can you show me some astrology symbols?''. Existing models tend to use rare Unicode characters or emojis to explain the symbols, but cannot understand the visual semantics embedded in those symbols flexibly.

\begin{figure*}[]
\centering
\subfigure[\small Text-only\label{fig:groups-llm}]{
\begin{minipage}[t]{\linewidth}
\centering
    \includegraphics[width=\linewidth]{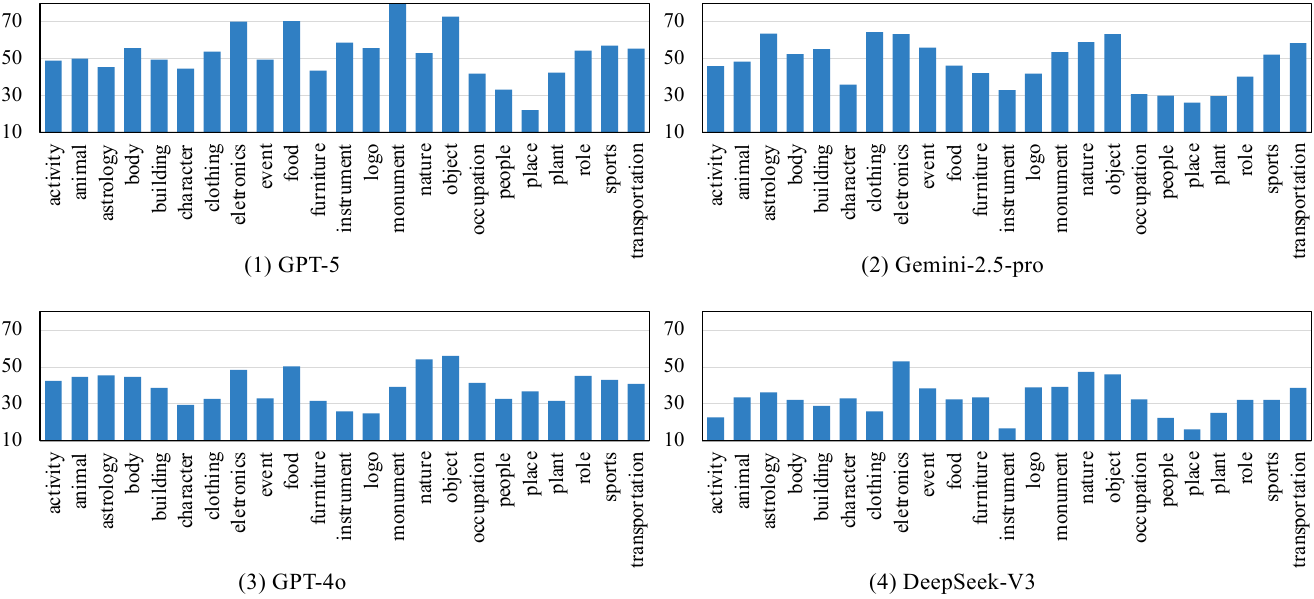}
    \includegraphics[width=\linewidth]{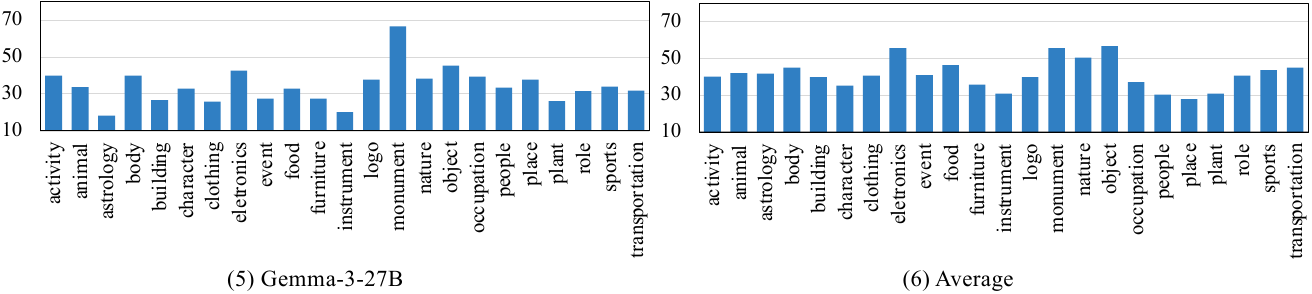}
\end{minipage}%
}%

\subfigure[\small{Image-only}\label{fig:groups-MLLM}]{
\begin{minipage}[t]{\linewidth}
\centering
    \includegraphics[width=\linewidth]{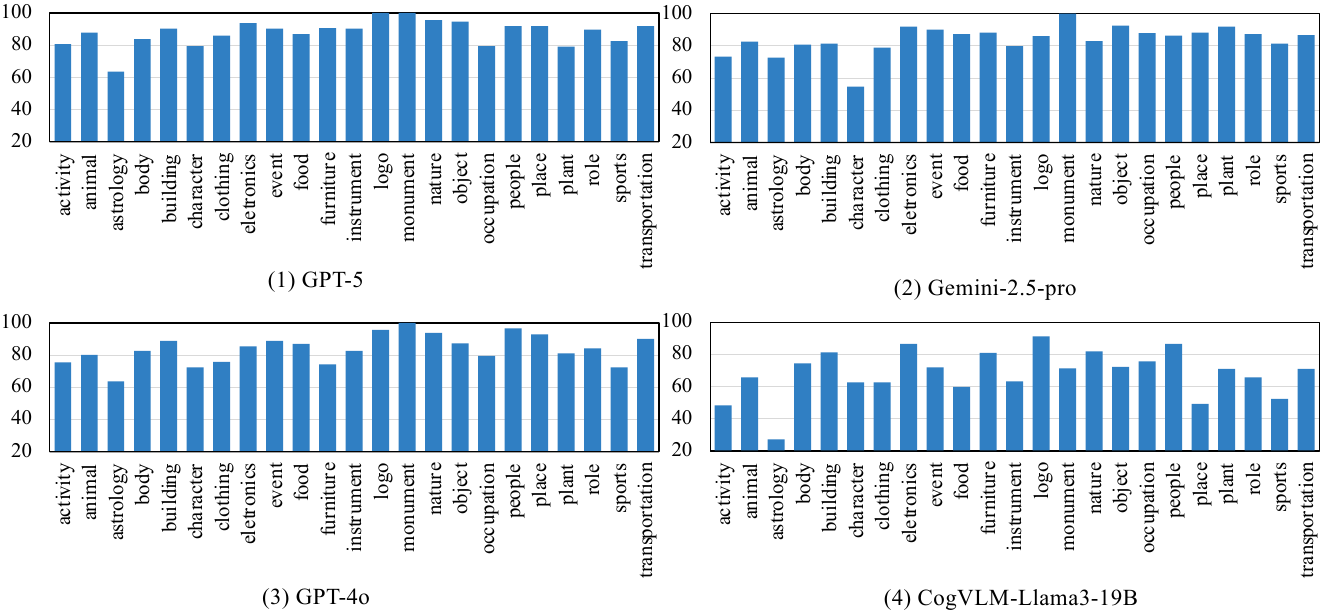}
    \includegraphics[width=\linewidth]{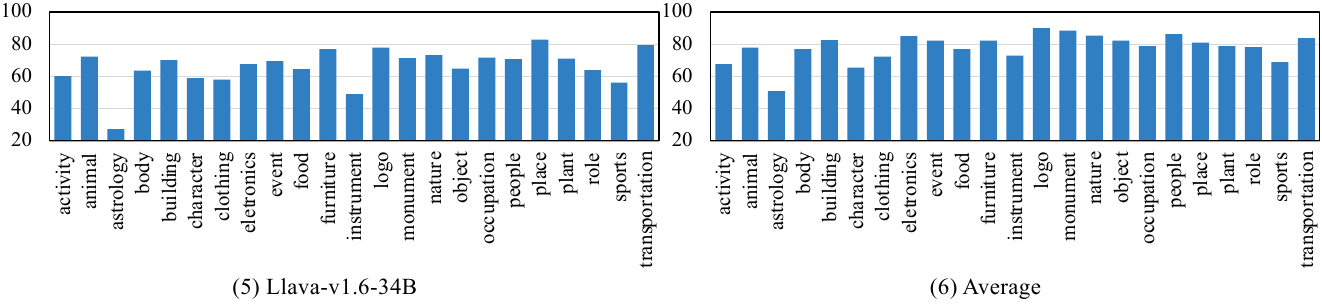}
\end{minipage}%
}%
\caption{Micro accuracy(\%) of models on recognizing ASCII arts in different groups. Average is calculated as the mean of the top 5 models.}
\label{fig:groups-performance}
\end{figure*}

The models' performance under different groups is shown in Fig.~\ref{fig:groups-performance}. Overall, the performance of models under Image-only mode is more balanced across different categories, except for the drops in ``astrology'' and ``character''. Meanwhile, accuracy given images fluctuates among different groups, with ``electronics'', ``monumnet'' and ``object'' topping the rank. 

\section{Case Studies}
\label{app:cases}

\begin{figure*}[h]
    \centering
    \includegraphics[width=0.95\linewidth]{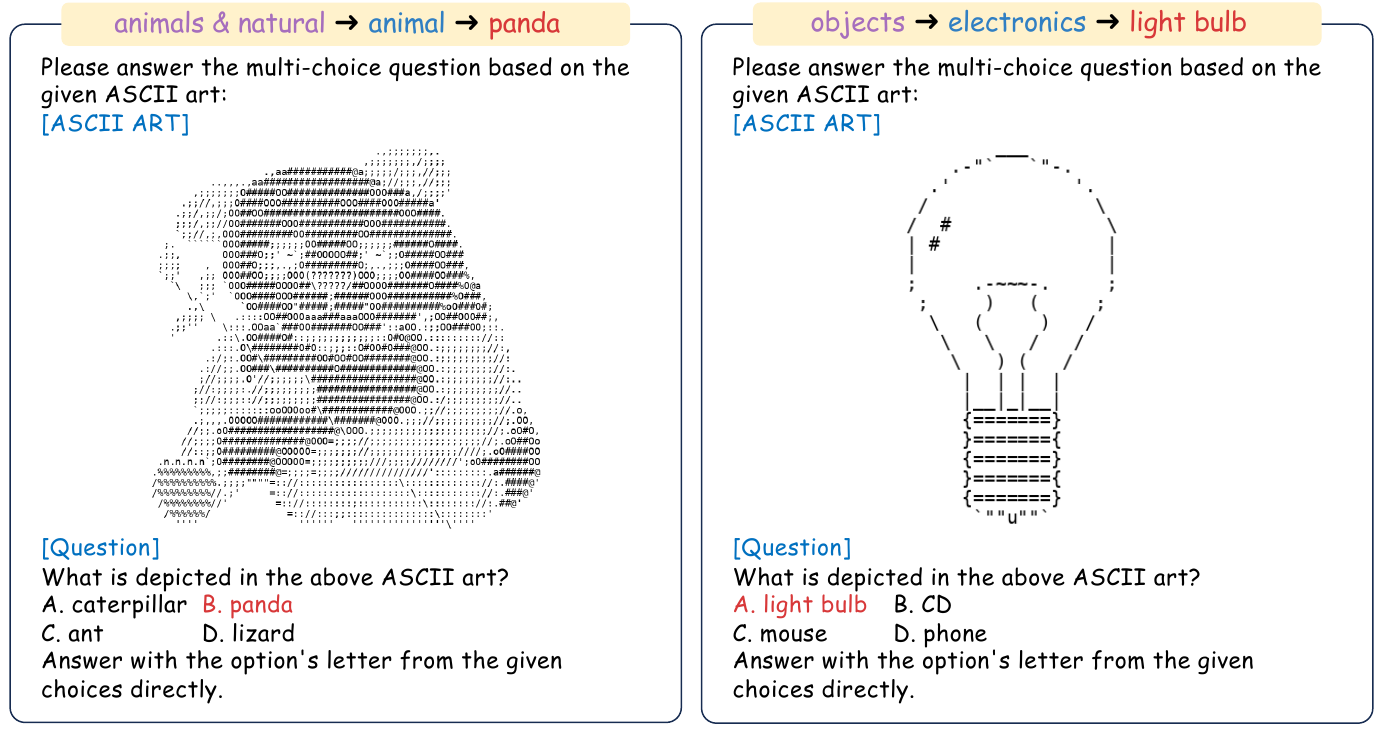}
    \includegraphics[width=0.95\linewidth]{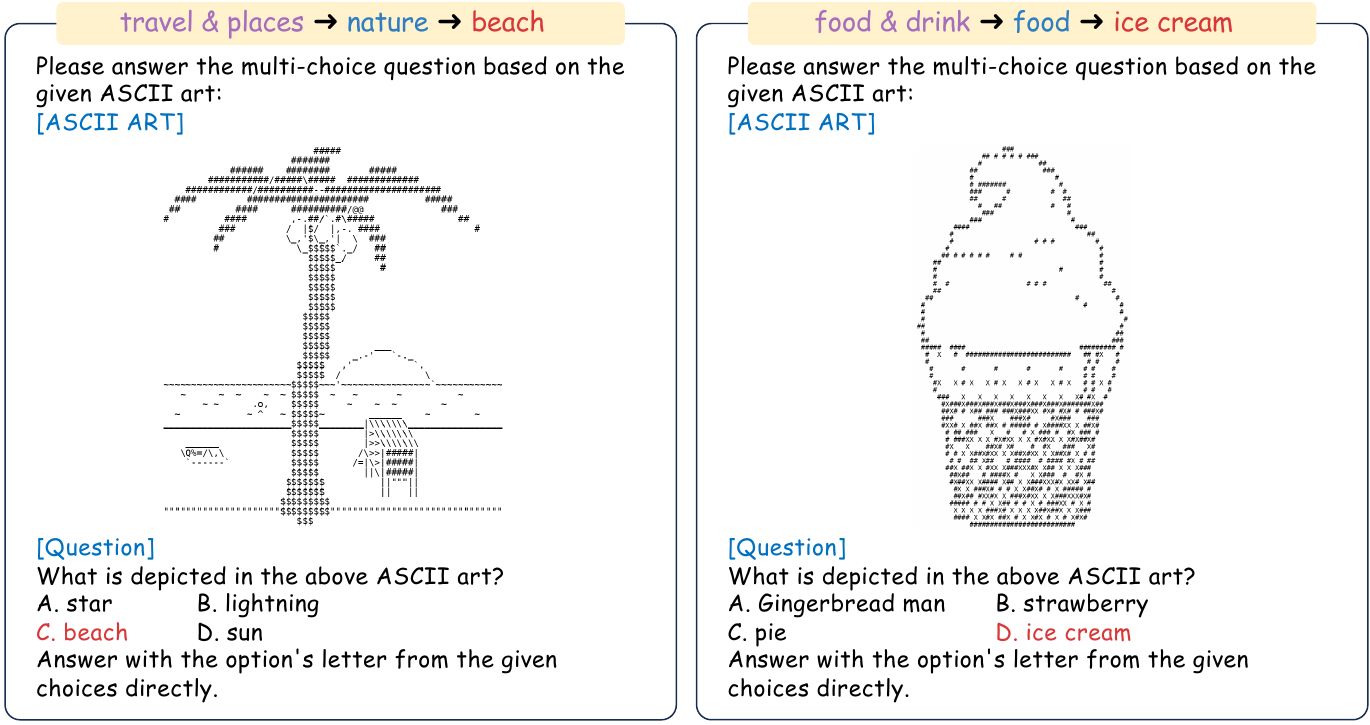}
    \caption{Case studies (Part I).}
    \label{fig:cases-1}
\end{figure*}

\begin{figure*}[h]
    \centering
    \includegraphics[width=0.95\linewidth]{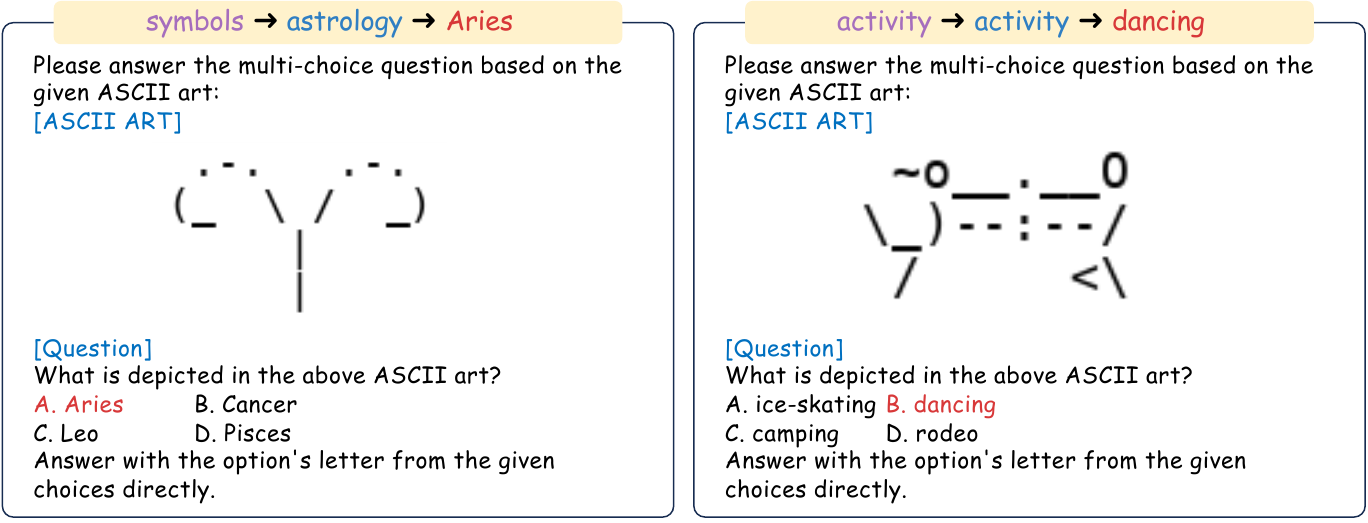}
    \includegraphics[width=0.475\linewidth]{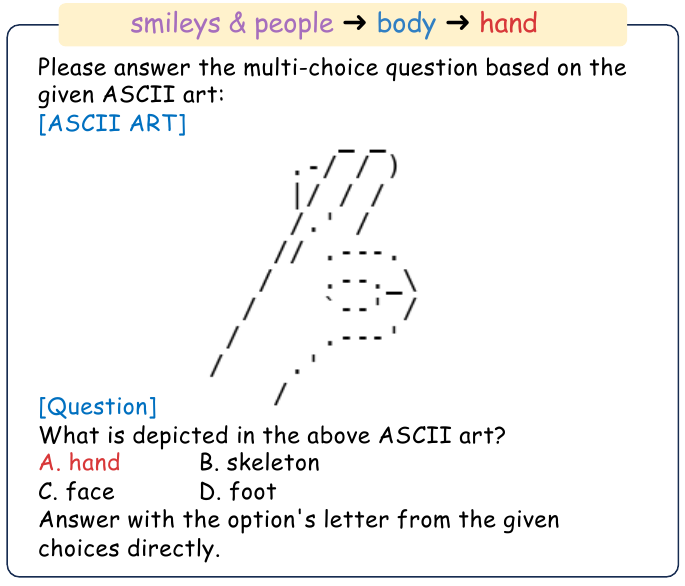}
    \caption{Case studies (Part II).}
    \label{fig:cases-2}
\end{figure*}

We selected seven samples belonging to different classes from \DataName{} and show the cases in Fig.~\ref{fig:cases-1} and Fig.~\ref{fig:cases-2}. The correct answers are marked in red.

\end{document}